\documentclass[11pt]{article}
\usepackage[utf8]{inputenc}
\usepackage{amsmath,amssymb,array}
\usepackage{graphicx}
\usepackage{booktabs}
\usepackage{multirow}
\usepackage{lscape}
\usepackage{setspace}
\usepackage{hyperref}
\usepackage{authblk}

\usepackage{geometry}
\usepackage{environ}
\usepackage[font=small,labelfont=bf]{caption}
\NewEnviron{widefigure}[1][]{
\begin{figure}[#1]
\advance\leftskip-2cm
\begin{minipage}{\dimexpr\textwidth+4cm\relax}%
  \captionsetup{margin=2cm}
  \BODY
\end{minipage}%
\end{figure}
}
\NewEnviron{widetable}[1][]{
\begin{table}[#1]
\advance\leftskip-2cm
\begin{minipage}{\dimexpr\textwidth+4cm\relax}%
  \captionsetup{margin=2cm}
  \BODY
\end{minipage}%
\end{table}
}

\usepackage{fancyvrb}
\usepackage{alltt}
\DefineVerbatimEnvironment{example}{Verbatim}{}
\renewenvironment{example*}{\begin{alltt}}{\end{alltt}}

\usepackage{natbib}
\title{\textbf{RFpredInterval: An R Package for Prediction Intervals with Random Forests and Boosted Forests} \vspace{0.6cm}}
\author{Cansu Alakuş\thanks{Corresponding author. E-mail: \href{mailto:cansu.alakus@hec.ca}{cansu.alakus@hec.ca}} }
\author{Denis Larocque}
\author{Aur\'elie Labbe}
\affil{Department of Decision Sciences, HEC Montr\'eal, Montr\'eal, QC H3T 2A7, Canada} 
\date{}

\begin{document}

\maketitle
\begin{abstract}
   Like many predictive models, random forests provide point predictions for new observations. Besides the point prediction, it is important to quantify the uncertainty in the prediction. Prediction intervals provide information about the reliability of the point predictions. We have developed a comprehensive R package, \texttt{RFpredInterval}, that integrates 16 methods to build prediction intervals with random forests and boosted forests. The set of methods implemented in the package includes a new method to build prediction intervals with boosted forests (PIBF) and 15 method variations to produce prediction intervals with random forests, as proposed by \cite{roy_prediction_2020}. We perform an extensive simulation study and apply real data analyses to compare the performance of the proposed method to ten existing methods for building prediction intervals with random forests. The results show that the proposed method is very competitive and, globally, outperforms competing methods.
\end{abstract}

\newpage
\section{Introduction}

Predictive modelling is the general concept of building a model that describes how a group of covariates can be used to predict a response variable. The objective is to predict the unknown responses of observations given their covariates. For example, predictive models could be used to predict the sale price of houses given house characteristics \citep{de_cock_ames_2011}. In its simplest form, a predictive model aims to provide a point prediction for a new observation. However, a point prediction does not contain information about its precision that can tell us how close to the true response we can expect the prediction to be, which is often important in decision-making context. Hence, although the point prediction is often the main goal of predictive analysis, assessing its reliability is equally important, and this can be achieved with a prediction interval (PI). A PI contains a set of likely values for the true response with an associated level of confidence, usually, $90\%$ or $95\%$. Given that shorter PIs are more informative, developing predictive models that can produce shorter PIs along with the point predictions is crucial in assessing and quantifying the prediction error. In real-world applications, knowing the prediction error alongside the point prediction increases the practical value of the prediction.

Regression analysis is a form of predictive modelling technique that examines the relationship between a response variable and a group of covariates. In this paper, we consider a general regression model
\begin{equation} \label{eq:regression}
Y = g\left(X\right)+\epsilon
\end{equation}
where $Y$ is a univariate continuous response variable, $X$ is a \textit{p}-dimensional vector of predictors, and $\epsilon$ is an error term. We assume $g\left(.\right)$ is an unknown smooth function $\Re^p \rightarrow \Re$ and $E\left[Y|X=x\right] = g\left(X\right)$. A confidence interval of the prediction is a range likely to contain the location of the response variable's true population mean. However, a prediction interval for a new observation is wider than its corresponding confidence interval and provides a range likely to contain this new observation's response value.

In the past decade, random forests have increased in popularity and provide an efficient way to generate point predictions for model \eqref{eq:regression}. A random forest is an ensemble method composed of many decision trees, which can be described with a simple algorithm \citep{breiman_random_2001}. For each tree $b=\left \{1,..., B\right \}$, a bootstrap sample of observations is drawn and a fully grown tree is built such that a set of predictors is randomly selected at each node and the best split is selected among all possible splits with those predictors only. The random forest prediction for a new observation is the average of the $B$ trees
\begin{equation*}
    \hat{y}^{}_{new} = \frac{1}{B} \sum_{b=1}^{B}\hat{y}^b_{new}
\end{equation*}
where $\hat{y}^b_{new}$ is the tree prediction for the new observation in the $b$th tree, i.e. the average of observations in the terminal node corresponding to the new observation. Besides this traditional description, the modern view also considers random forests as data-driven weight generators  \citep{hothorn_bagging_2004,lin_random_2006,moradian_l1_2017,moradian_survival_2019,athey_generalized_2019,roy_prediction_2020,tabib_non-parametric_2020,alakus_conditional_2021}. 

Although random forests limit over-fitting by combining many trees, which reduces the variance of the estimator, final predictions can be biased \citep{mentch_quantifying_2016,wager_estimation_2018}. Since each tree is built under the same random process, all trees focus on the same part of the response signal, usually the strongest. Therefore, some parts of the response signal may be left untargeted, which could result in biased point predictions. \citet{wager_estimation_2018} provide bounds for the extent of the bias of random forests under some assumptions about the tree growing process. Following their work, \citet{ghosal_boosting_2021} proposed a bias correction method in a regression framework called \textit{one-step boosted forest}, which is introduced in \citet{breiman_random_2001} and \citet{zhang_bias-corrected_2012}. The main idea of the proposed method is to sum the predictions of two random forests, where the first is a regression forest fitted on the target data set and the second is fitted on the out-of-bag residuals of the former. Empirical studies show that this method provides a significant bias reduction when compared to a simple random forest. 

The current paper proposes an R package providing, among other features, an extension of the one-step boosted forest method described above \citep{ghosal_boosting_2021}. The literature on prediction intervals for random forests consists mostly of recent studies. The first method is the Quantile Regression Forests (QRF) method proposed by \citet{meinshausen_quantile_2006}. The aim of QRF is to estimate conditional quantiles of the response variable, instead of conditional means, using an estimated cumulative distribution function obtained with the nearest neighbour forest weights introduced by \citet{hothorn_bagging_2004}. Prediction intervals can be built directly from the estimated conditional quantiles. The method is implemented in the CRAN package \texttt{quantregForest} \citep{R-quantregForest}.

In a more recent study, \citet{athey_generalized_2019} proposed Generalized Random Forests (GRF), a very general framework to estimate any quantity, such as conditional means, quantiles or average partial effects, identified by local moment equations. Trees are grown with splitting rules designed to maximize heterogeneity with respect to the quantity of interest. Quantile regression forest is one of the applications of GRF. Similar to the QRF, the GRF method uses the neighbourhood information from different trees to compute a weighted set of neighbours for each test point. Unlike QRF, which grows trees with the least-squares criterion, GRF uses a splitting rule designed to capture heterogeneity in conditional quantiles. An implementation of quantile regression forest with GRF is available in the function \texttt{quantile\_forest} of the CRAN package \texttt{grf} \citep{R-grf}.

\citet{vovk_algorithmic_2005,vovk_-line_2009} introduced a general distribution-free conformal prediction interval framework. Any predictive model, including random forests, can be used within the proposed methodology. The idea is to use an augmented data set that includes the new observation to be predicted to fit the model, and apply a set of hypothesis tests to provide an error bound around the point prediction for the new observation. Although this method does not require any distribution assumptions, it is computationally intensive. \citet{lei_distribution-free_2018} proposed a variant of this method, called Split Conformal (SC) prediction, which splits the data into two subsets, one to fit the model, and one to compute the quantiles of the residual distribution. We note that, while the original full conformal prediction interval framework produces shorter intervals, SC is computationally more efficient. The R package \texttt{conformalInference} \citep{R-conformalInference}, available on GitHub, implements this method.

\citet{roy_prediction_2020} proposed 20 distinct variations of methods to improve the performance of prediction intervals with random forests. These approaches differ according to 1) the method used to build the forest and 2) the method used to build the prediction interval. Four methods can be used to build the forest: three from the classification and regression tree (CART) paradigm \citep{breiman_classification_1984} and the transformation forest method (TRF) proposed by \cite{hothorn_predictive_2021}. Within the CART paradigm, in addition to the default least-squares (LS) splitting criterion, two alternative splitting criteria, $L_1$ and shortest prediction interval (SPI), are considered. Prediction intervals are built using the Bag of Observations for Prediction (BOP), which is the set of nearest neighbour observations previously used in \citet{moradian_l1_2017,moradian_survival_2019}. In addition to the type of forest chosen, there are also five methods to build prediction intervals: the classical method (LM), the quantile method (Quant), the shortest prediction interval (SPI), the highest density region (HDR), and the contiguous HDR (CHDR). LM is computed based on an intercept-only linear model using the BOP as the sample, and produces a symmetric PI around the point prediction. The quantile method, similar to the QRF method, is based on the quantiles of the BOP. SPI corresponds to the shortest interval among the intervals that contain at least $\left(1-\alpha\right)100\%$ of the observations in the BOP. As an alternative to SPI, HDR is the smallest region in the BOP, with the desired coverage $\left(1-\alpha\right)$. Note that HDR is not necessarily a single interval. If the distribution is multimodal, it can be formed by multiple intervals. Finally, CHDR is a way to obtain a single prediction interval from HDR intervals by building an interval with the minimum and maximum bounds of the HDR intervals. 

\citet{zhang_random_2020} proposed a forest-based prediction interval method, called Out-of-Bag (OOB) prediction intervals, to estimate prediction intervals using the empirical quantiles of the out-of-bag prediction errors. The method assumes that OOB prediction errors are identically distributed and that their distribution can be well approximated by the out-of-bag prediction errors obtained from all training observations. The resulting prediction intervals have the same width for all test observations. The method is implemented in the CRAN package \texttt{rfinterval} \citep{R-rfinterval}.

\citet{lu_unified_2021} proposed a very interesting and useful method to estimate the conditional prediction error distribution of a random forest. The main idea of the proposed method is to use a random forest to compute the out-of-bag residuals for the training observations and to form a set of out-of-bag neighbours for each test point. Then, the conditional prediction error distribution for each test point is determined with the out-of-bag residuals in the neighbourhood. Estimating the prediction error distribution enables the estimation of conditional quantiles, conditional biases and conditional mean squared prediction errors. The prediction interval for a test point $x$, defined by $\widehat {PI}_{\alpha}\left(x\right)$, is formed by adding the $\alpha/2$ and $1-\alpha/2$ quantiles of the conditional prediction error distribution to the random forest point prediction. The estimators are implemented in the CRAN package \texttt{forestError} \citep{R-forestError}. 

Note that the conformal inference, OOB approach of \citet{zhang_random_2020} and the $\widehat {PI}_{\alpha}\left(x\right)$ method of \cite{lu_unified_2021} all use the prediction errors to build the prediction intervals. Instead of using the training responses directly to estimate quantiles, using prediction errors provides a better predictive power. However, unlike conformal inference and the OOB approach, the $\widehat {PI}_{\alpha}\left(x\right)$ method uses the nearest neighbour observations to estimate the prediction error distribution. This idea is very similar to the BOP idea \citep{roy_prediction_2020}, but instead of using in-bag observations, \citet{lu_unified_2021} use out-of-bag observations to form the neighbourhoods. This approach allows the local information for the test observations to be extracted. 
	
In this paper, we introduce the R package \texttt{RFpredInterval} \citep{R-RFpredInterval}, which is the novel implementation of 16 methods to build prediction intervals with random forests and boosted forests. The set of methods implemented in the package includes a new method to build prediction intervals with boosted forests and 15 method variations (three splitting rules with the CART paradigm which are LS, $L_1$  and SPI, and five methods to build prediction intervals which are LM, SPI, Quant, HDR and CHDR) proposed by \citet{roy_prediction_2020}. These 15 methods had been thoroughly investigated before through simulation studies and with real data sets in \citet{roy_prediction_2020}. However, they are not easily available to use. One of the main contributions of our package is the implementation of these competitive methods and the ability for users to compare various prediction interval methods within the same package. The other main contribution of our paper is a new method to build prediction intervals. Contrary to the 15 methods proposed by \citet{roy_prediction_2020}, the newly introduced method was not tested before. That is why in the paper we placed a greater emphasis on investigating the new method through extensive simulation studies and with real data. For performance comparison purposes, we compared the new method to 10 existing methods which include: 
\begin{itemize}
	\item 5 of the 15 implemented method variations of \citet{roy_prediction_2020}; see the Competing methods subsection for details.
    \item 5 other competing methods from the literature: Quantile Regression Forests (QRF), Generalized Random Forests (GRF), Split Conformal prediction method (SC), Out-of-Bag (OOB) prediction intervals method and $\widehat {PI}_{\alpha}\left(x\right)$ method.
\end{itemize}

The new proposed method to build \textbf{P}rediction \textbf{I}ntervals with \textbf{B}oosted \textbf{F}orests is called \textbf{PIBF}. This approach integrates the idea of using the nearest neighbour out-of-bag observations to estimate the conditional prediction error distribution presented in \citet{lu_unified_2021} to the one-step boosted forest proposed by \citet{ghosal_boosting_2021}. We will show in this paper that PIBF significantly improves the performance of prediction intervals with random forests when compared with 10 existing methods using a variety of simulated and real benchmark data sets. 
 
The rest of the paper is organized as follows. In the next section, we describe the algorithm implemented in PIBF. We then present the details of the package and provide a practical and reproducible example. We also perform a simulation study to compare the performance of our proposed method to existing competing methods, and we investigate the performance of the proposed method with real data sets. Lastly, we conclude with a discussion of the results.

\section{Method and implementation}

The proposed method is based on the one-step boosted forest method proposed by \citet{ghosal_boosting_2021}. It consists in fitting two regression random forests: the first is fitted to get point predictions and out-of-bag (OOB) residuals using the given data set, whereas the second is fitted to predict those residuals using the original covariates. As empirical studies demonstrate, the one-step boosted forest provides point predictions with reduced bias compared to the simple random forest. \citet{ghosal_boosting_2021} use subsampling for their theoretical investigations, even though random forests were originally described with bootstrap samples and obtain notable performance improvements. They have also investigated the effect of using bootstrapping with the one-step boosted forest on the bias estimations. From the results presented in their appendix, the use of bootstrapping yields the best performance and reduces the bias the most in exchange for an increase in their proposed variance estimator, which is defined under asymptotic normality. In this paper, we use bootstrapping for the one-step boosted forest method following the better performance results on bias reduction. The final prediction for a new observation, $x_{new}$, is the sum of the predictions from the two random forests
\begin{equation}
    \hat y^{*}_{new} = \hat y^{}_{new} + \hat \epsilon^{}_{new}
\end{equation}
where $\hat y^{}_{new}$ is the point prediction obtained from the first random forest and $\hat \epsilon^{}_{new}$ is the bias estimation from the second forest.

Besides bias correction, we use the second random forest as a way to construct a prediction interval by finding the nearest neighbour observations that are close to the one we want to predict. The idea of finding the nearest neighbour observations, a concept very similar to the 'nearest neighbour forest weights' \citep{hothorn_bagging_2004, lin_random_2006}, was introduced in \cite{moradian_l1_2017} and later used in \cite{moradian_survival_2019}, \cite{roy_prediction_2020}, \cite{tabib_non-parametric_2020} and \cite{alakus_conditional_2021}. For a new observation, the set of in-bag training observations that are in the same terminal nodes as the new observation forms the set of nearest neighbour observations. \cite{roy_prediction_2020} called this set of observations the Bag of Observations for Prediction (BOP). We can define the BOP for a new observation $x_{new}$ as
\begin{equation} \label{eq:bop}
    BOP\left(x_{new}\right) = \bigcup\limits_{b=1}^{B} I_b\left(x_{new}\right)
\end{equation}
where $I_b\left(x_{new}\right)$ is the set of in-bag training observations, \textit{i.e.}, observations in the bootstrap sample that are in the same terminal node as $x_{new}$ in the $b^{th}$ tree. $I_b\left(.\right)$ consists of the training observations that are in the bootstrap sample of the $b^{th}$ tree.

Instead of forming the set of nearest neighbour observations with the in-bag training observations, we can use the out-of-bag observations which are not in the bootstrap sample, as used in \citet{lu_unified_2021}. We can define the out-of-bag equivalent of the BOP for a new observation $x_{new}$ \eqref{eq:bop} as 
\begin{equation} \label{eq:bop2}
    BOP^{*}\left(x_{new}\right) = \bigcup\limits_{b=1}^{B} O_b\left(x_{new}\right)
\end{equation}
where $O_b\left(x_{new}\right)$ is the set of out-of-bag observations that are in the same terminal node as $x_{new}$ in the $b^{th}$ tree. $O_b\left(.\right)$ consists of the training observations that are not in the bootstrap sample of the $b^{th}$ tree. 

Out-of-bag observations are not used in the tree growing process. Thus, for the trees where the training observations are out-of-bag, they are like the unobserved test observations for those trees. The only difference is that, for a new observation, we use all the trees in the forest whereas for an out-of-bag observation we have only a subset of the forest trees. By using the out-of-bag equivalent of the BOP for a new observation, we can make use of the analogy between the out-of-bag observations and test observations. The out-of-bag neighbours of a new observation represent the new observation better than the in-bag neighbours.

Any desired measure can be obtained by using the constructed BOPs. In this paper, we use the BOP idea to build a prediction interval for a test observation. For a new observation with covariates $x_{new}$, we firstly form $BOP^{*}\left(x_{new}\right)$ \eqref{eq:bop2} using the out-of-bag neighbours. Then, as proposed by \citet{lu_unified_2021}, we estimate the conditional prediction error distribution, $\hat{F}\left(x_{new}\right)$, but now with the bias-corrected out-of-bag residuals of the observations in $BOP^{*}\left(x_{new}\right)$. Lastly, we build a prediction interval for the new observation as
\begin{equation}
    PI\left(x_{new}\right) = \left[\hat y^{*}_{new} + SPI^l_{\alpha}\left(\hat{F}\left(x_{new}\right)\right), \hat y^{*}_{new} + SPI^u_{\alpha}\left(\hat{F}\left(x_{new}\right)\right)\right]
\end{equation}
where $\hat y^{*}_{new}$ is the bias-corrected prediction, $SPI^l_{\alpha}\left(\hat{F}\left(x_{new}\right)\right)$ and $SPI^u_{\alpha}\left(\hat{F}\left(x_{new}\right)\right)$ are the lower and upper bounds of the $SPI_{\alpha}\left(\hat{F}\left(x_{new}\right)\right)$, which is the shortest interval formed by the observations in $BOP^{*}\left(x_{new}\right)$ that contains at least $\left(1-\alpha\right)100\%$ of the observations. By using the bias-corrected residuals to form prediction error distribution and picking the shortest interval among the qualified intervals, we can expect narrower prediction intervals. 

We can summarize the steps of the proposed method as follows:
\begin{enumerate}
    \item Train the first regression RF with covariates $X$ to predict the response variable $Y$, and get the OOB predictions $\hat Y_{oob}$
    \item Compute the OOB residuals as $\hat \epsilon_{oob}^{} = Y-\hat Y_{oob}^{}$
    \item Train the second regression RF with covariates $X$ to predict the OOB residuals $\hat \epsilon_{oob}^{}$, and get the OOB predictions for residuals $\hat{\hat \epsilon}_{oob}^{}$
    \item Update the OOB predictions as $\hat Y_{oob}^* = \hat Y_{oob}^{} + \hat{\hat \epsilon}_{oob}^{}$
    \item Compute the updated OOB residuals after bias-correction as $\hat \epsilon_{oob}^{*} = Y - \hat Y_{oob}^*$
    \item For a new observation $x_{new}$, get the point predictions $\hat y_{new}$ from the first RF, and get the predicted residuals $\hat \epsilon_{new}$ from the second RF, then the final prediction for the new observation is $$\hat y_{new}^{*} = \hat y_{new}^{} + \hat \epsilon_{new}^{}$$ where $\hat \epsilon_{new}^{}$ is the estimated bias.
    \item Form a BOP for $x_{new}$ with the OOB neighbours using the second RF, $BOP^{*}\left(x_{new}\right) \eqref{eq:bop2}$ and estimate the conditional prediction error distribution for $x_{new}$ as, $$\hat{F}\left(x_{new}^{}\right) = \left \{\hat \epsilon_{oob,i}^{*} | i \in BOP^{*}\left(x_{new}\right)\right \}$$
    \item Build a PI for $x_{new}$ as $PI\left(x_{new}\right) = \left[\hat y^{*}_{new} + SPI^l_{\alpha}\left(\hat{F}\left(x_{new}\right)\right), \hat y^{*}_{new} + SPI^u_{\alpha}\left(\hat{F}\left(x_{new}\right)\right)\right]$
\end{enumerate}

\subsection{Calibration}
The principal goal of any prediction interval method is to ensure the desired coverage level. In order to attain the desired coverage level $\left(1-\alpha\right)$, we may need a calibration procedure. The goal of the calibration is to find the value of $\alpha_w$, called the working level in \citet{roy_prediction_2020}, such that the coverage level of the PIs for the training observations is closest to the desired coverage level. \citet{roy_prediction_2020} presented a calibration procedure that uses the BOPs that are built using only the trees where the training observation $x_i$ is OOB. The idea is to find the value of $\alpha_w$ using the OOB-BOPs. In this paper, we call this procedure OOB calibration. 

We also include a cross-validation-based calibration procedure with the proposed method to acquire the desired $\left(1-\alpha\right)$ coverage level. In this calibration, we apply $k$-fold cross-validation to form prediction intervals for the training observations. In each fold, we split the original training data set into training and testing sets. For the training set, we go through the steps 1-5 defined above. Then, for each observation in the testing set, we apply steps 6-8 and build a PI. After completing CV, we compute the coverage level with the constructed PIs and if the coverage is not within the acceptable coverage range, then we apply a grid search to find the $\alpha_w$ such that $\alpha_w$ is the closest to the target $\alpha$ among the set of $\alpha_w$'s. Once we find the $\alpha_w$, we use this level to build the PI for the new observations. 

\subsection{The \texttt{RFpredInterval} package}

In our package, we implement 16 methods that apply random forest training. Ten of these methods have specialized splitting rules in the random forest growing process. These methods are the ones with $L_1$ and shortest prediction interval (SPI) splitting rules proposed by \citet{roy_prediction_2020}. To implement these methods, we have utilised the custom split feature of the \texttt{randomForestSRC} package \citep{R-randomForestSRC}.

The \texttt{randomForestSRC} package allows users to define a custom splitting rule for the tree growing process. The user needs to define the customized splitting rule in the \texttt{splitCustom.c} file with C-programming. After modifying the \texttt{splitCustom.c} file, all C source code files in the package's \texttt{src} folder must be recompiled. Finally, the package must be re-installed for the custom split rule to become available.

In our package development process, we froze the version of \texttt{randomForestSRC} to the latest one available at the time, which is version 2.11.0, to apply specialized splitting rules. After defining the $L_1$ and SPI splitting rules, all C files were re-compiled. Finally, all package files including our R files for prediction interval methods were re-built to make the package ready for the user installation.

The \texttt{RFpredInterval} package has two main R functions as below:
\begin{itemize}
    \item \texttt{pibf()}: Constructs prediction intervals with the proposed method, PIBF.
    \item \texttt{rfpi()}: Constructs prediction intervals with 15 distinct variations proposed by \citet{roy_prediction_2020}. 
\end{itemize}

Table~\ref{table:methods} presents the list of functions and methods implemented in \texttt{RFpredInterval}. For \texttt{pibf()}, \texttt{RFpredInterval} uses the CRAN package \texttt{ranger} \citep{R-ranger} to fit the random forests. For \texttt{rfpi()}, \texttt{RFpredInterval} uses \texttt{randomForestSRC} package. For the least-squares splitting rule, both \texttt{randomForestSRC} and \texttt{ranger} packages are applicable.

\begin{widetable}[t]
\caption{\label{table:methods}List of functions and methods with characteristics}
\centering
\resizebox{\linewidth}{!}{
\begin{tabular}[t]{lll>{\raggedright\arraybackslash}p{18.5cm}}
\toprule
\multicolumn{1}{l}{\textbf{Function}} & \multicolumn{2}{l}{\textbf{Method}} & \multicolumn{1}{l}{\textbf{Details}}\\
\midrule
\texttt{pibf()} & PIBF &  & Builds prediction intervals with boosted forests. The \texttt{ranger} package is used to fit the random forests. Calibration options are \texttt{"cv"} and \texttt{"oob"}. Returns constructed PIs and bias-corrected point predictions for the test data.\\
\cmidrule{1-4}
 & \textit{Split method} & \textit{PI method} & \\
\cmidrule{2-3}
 &  & LM & \\
 &  & SPI & \\
 &  & Quant & \\
 &  & HDR & \\
 & \multirow[t]{-5}{*}{\raggedright\arraybackslash LS} & CHDR & \multirow[t]{-5}{18.5cm}{\raggedright\arraybackslash Splitting rule is the least-squares (LS) from the CART paradigm. \texttt{"ls"} is used for the \texttt{"split\_method"} argument. A vector of characters \texttt{c("lm","spi","quant","hdr","chdr")} is used for the \texttt{"pi\_method"} argument to apply all or a subset of the PI methods. \texttt{ranger} or \texttt{randomForestSRC} can be used to fit the random forest. Returns a list of constructed PIs for the selected PI methods and point predictions for the test data.}\\
\cmidrule{2-4}
 &  & LM & \\
 &  & SPI & \\
 &  & Quant & \\
 &  & HDR & \\
 & \multirow[t]{-5}{*}{\raggedright\arraybackslash $L_1$} & CHDR & \multirow[t]{-5}{18.5cm}{\raggedright\arraybackslash Splitting rule is the $L_1$ from the CART paradigm \citep{roy_prediction_2020}. \texttt{"l1"} is used for the \texttt{"split\_method"} argument. A vector of characters \texttt{c("lm","spi","quant","hdr","chdr")} is used for the \texttt{"pi\_method"} argument to apply all or a subset of the PI methods. Only \texttt{randomForestSRC} can be used to fit the random forest since the split rule is implemented with the custom split feature of that package. Returns a list of constructed PIs for the selected PI methods and point predictions for the test data.}\\
\cmidrule{2-4}
 &  & LM & \\
 &  & SPI & \\
 &  & Quant & \\
 &  & HDR & \\
\multirow[t]{-16}{*}{\raggedright\arraybackslash \texttt{rfpi()}} & \multirow[t]{-5}{*}{\raggedright\arraybackslash SPI} & CHDR & \multirow[t]{-5}{18.5cm}{\raggedright\arraybackslash Splitting rule is the shortest PI (SPI) from the CART paradigm \citep{roy_prediction_2020}. \texttt{"spi"} is used for the \texttt{"split\_method"} argument. A vector of characters \texttt{c("lm","spi","quant","hdr","chdr")} is used for the \texttt{"pi\_method"} argument to apply all or a subset of the PI methods. Only \texttt{randomForestSRC} can be used to fit the random forest since the split rule is implemented with the custom split feature of that package. Returns a list of constructed PIs for the selected PI methods and point predictions for the test data.}\\
\cmidrule{1-4}
\texttt{piall()} & All methods &  & Builds prediction intervals with all of the implemented PI methods. The \texttt{ranger} package is used to fit the random forests for the PIBF and methods with LS split rule. \texttt{randomForestSRC} package is used for the methods with $L_1$ and SPI split rules. Default values are assigned to the function arguments of \texttt{pibf()} and \texttt{rfpi()}. Returns an object of class \texttt{"piall"} containing a list of constructed PIs with 16 methods, and point predictions obtained with the PIBF method, LS, $L_1$ and SPI split rules for the test data.\\
\cmidrule{1-4}
\texttt{plot()} &  &  & Plots the 16 constructed PIs obtained with \texttt{piall()} function for a test observation. \\
\cmidrule{1-4}
\texttt{print()} &  &  & Prints the summary output of \texttt{pibf()}, \texttt{rfpi()} and \texttt{piall()} functions. \\
\bottomrule
\multicolumn{4}{l}{\textbf{LM}: Classical method, \textbf{SPI}: Shortest PI, \textbf{Quant}: Quantiles, \textbf{HDR}: Highest density region, \textbf{CHDR}: Contiguous HDR}\\
\end{tabular}}
\end{widetable}

In this section, we illustrate the usage of the \texttt{RFpredInterval} package with the Ames Housing data set \citep{de_cock_ames_2011}. The data set was introduced as a modern alternative to the well-known Boston Housing data set. The data set contains many explanatory variables on the quality and quantity of physical attributes of houses in Ames, IA sold from 2006 to 2010. Most of the variables give information to a typical home buyer who would like to know about a house (\textit{e.g.} number of bedrooms and bathrooms, square footage, heating type, lot size, etc.).

The \texttt{AmesHousing} \citep{R-AmesHousing} package contains the raw data and processed versions of the Ames Housing data set. The raw data contains 2930 observations and 82 variables, which include 23 nominal, 23 ordinal, 14 discrete, and 20 continuous variables, involved in assessing house values. The processed version of the data set has 2330 observations and 81 variables, including the target variable \texttt{Sale\_Price} representing the value of houses in US dollars. The usual goal for this data set is to predict the sale price of each house given covariates.

We load the processed version of the Ames Housing data set from the \texttt{AmesHousing} package and prepare the data set that we will use for the analyses. The preprocessing steps are presented in the Supplementary Material. This version of the data set contains 22 factors and 59 numeric variables, including 1 response variable \texttt{Sale\_Price}, for 2929 observations. We split the data set into training and testing samples.

\begin{example*}
set.seed(3456)
n <- nrow(AmesHousing)
trainindex <- sample(1:n, size = round(0.7*n), replace = FALSE)
traindata <- AmesHousing[trainindex, ]
testdata <- AmesHousing[-trainindex, ]
\end{example*}

We fit a random forest with 1000 trees using the training data and construct $95\%$ prediction intervals for the observations in the testing data with the proposed method. We apply 5-fold cross-validation based calibration and set the acceptable coverage range to $\left[.945,.955\right]$. We can pass the list of random forest parameters for \texttt{ranger} package.

\begin{example*}
out <- pibf(formula = Sale_Price ~ .,
            traindata = traindata, 
            testdata = testdata,
            alpha = 0.05,
            calibration = "cv", 
            numfolds = 5,
            coverage_range = c(0.945, 0.955),
            params_ranger = list(num.trees = 1000),
            oob = TRUE)
\end{example*}

We can then analyze the constructed PIs and bias-corrected random forest predictions for the testing data, as shown below. The PI output is a list containing lower and upper bounds. For example, we can print the point prediction and prediction interval for the tenth observation in the testing data.

\begin{example*}
out$pred_interval
out$test_pred
c(out$pred_interval$lower[10], out$test_pred[10], out$pred_interval$upper[10])
[1] 133.8629 160.2426 194.5804
\end{example*}

We can also print the summary output. In the summary output, we can always see the mean PI length over the test data set. If calibration is applied, we can see the working level of $\alpha$. If the test data set has true response information, as in our example, coverage and prediction errors for the test set are also printed. Moreover, since we have entered \texttt{oob = TRUE} in the function arguments, in the summary output we can see the mean PI length and coverage measures along with the prediction errors for the training set. The prediction intervals are built with the out-of-bag (OOB) predictions and prediction errors.

\begin{example*}
print(out)

>                       alpha_w: 0.050
>                Mean PI length: 73.081
>                      Coverage: 96.8%
>       MAE of test predictions: 12.773
>      RMSE of test predictions: 19.545
> 
>      Mean PI length (OOB PIs): 74.823
>            Coverage (OOB PIs): 94.7%
>  MAE of OOB train predictions: 14.179
> RMSE of OOB train predictions: 23.875
\end{example*}

Next, we construct $95\%$ prediction intervals using the variations proposed by \citet{roy_prediction_2020}. In the following example, the splitting is rule is set to $L_1$ and we want to apply LM, Quant and SPI methods for building prediction intervals. We apply OOB calibration and set the acceptable coverage range to $\left[.945,.955\right]$. We can pass the the list of random forest parameters for \texttt{randomForestSRC} package.

\begin{example*}
out2 <- rfpi(formula = Sale_Price ~ .,
             traindata = traindata, 
             testdata = testdata, 
             alpha = 0.05,
             calibration = TRUE, 
             split_rule = "l1", 
             pi_method = c("lm", "quant", "spi"),
             params_rfsrc = list(ntree = 1000),
             params_calib = list(range = c(0.945, 0.955)),
             oob = FALSE)
\end{example*}

We can analyze the constructed PIs for the testing data as below. Each PI output is a list containing lower and upper bounds. For instance, we can print the point prediction and LM prediction interval for the tenth observation in the testing data.

\begin{example*}
out2$lm_interval
out2$quant_interval
out2$spi_interval
c(out2$lm_interval$lower[10], out2$test_pred[10], out2$lm_interval$upper[10])
[1] 129.9474 154.2098 176.8429
\end{example*}

We print the summary output. In the summary output, we can see the splitting rule selected in the first row. Since the test data set has true responses in our example, we can see the coverage information for the selected PI methods besides the mean PI length and $\alpha_w$ in the printed table. Below the table, we have the mean prediction errors for the test set.

\begin{example*}
print(out2)

>                    Split rule: L1
> ------------------------------------------------------------------------------
>                                       Mean PI length    Coverage     alpha_w
> Classical method (LM)                     81.641          96.2%       0.140
> Shortest prediction interval (SPI)        81.953          96.1%       0.100
> Quantile method (Quant)                   80.893          95.8%       0.120
> ------------------------------------------------------------------------------
>       MAE of test predictions: 14.605
>      RMSE of test predictions: 22.603
\end{example*}

Although, with the \texttt{pibf()} and \texttt{rfpi()} functions, we have more flexibility to set the arguments for the methods, we can build prediction intervals with all 16 methods implemented in the package with the \texttt{piall()} function. We will build 95\% prediction intervals for the test set.

\begin{example*}
out3 <- piall(formula = Sale_Price ~ .,
              traindata = traindata, 
              testdata = testdata, 
              alpha = 0.05,
              num.trees = 1000)
\end{example*}

The output is a list of constructed prediction intervals with 16 methods and point predictions obtained with the PIBF method, LS, $L_1$, and SPI split rules. Hence, the output includes 16 prediction intervals and 4 point predictions which is a list of 20 items in total. 

We print the summary output.

\begin{example*}
print(out3)

> ---------------------------------------- 
>              Mean PI length    Coverage
> PIBF             72.752          96.6%
> LS-LM            81.800          96.5%
> LS-SPI           82.662          95.4%
> LS-Quant         81.523          95.2%
> LS-HDR           81.733          96.0%
> LS-CHDR          83.025          96.1%
> L1-LM            81.649          96.1%
> L1-SPI           81.805          96.1%
> L1-Quant         80.836          95.3%
> L1-HDR           83.032          96.4%
> L1-CHDR          82.482          96.1%
> SPI-LM           81.578          96.0%
> SPI-SPI          81.960          96.2%
> SPI-Quant        81.036          95.4%
> SPI-HDR          84.404          96.6%
> SPI-CHDR         82.927          96.1%
> ---------------------------------------- 
>                    MAE            RMSE
> PIBF             12.774         19.428
> LS split         14.412         22.413
> L1 split         14.596         22.569
> SPI split        14.558         22.600
\end{example*}

Lastly, we plot the constructed prediction intervals with all 16 methods, for the 15th observation in the test set.

\begin{example*}
plot(out3, test_id = 15)
\end{example*}

\begin{figure}[htbp]
  \centering
  \includegraphics[scale=0.7]{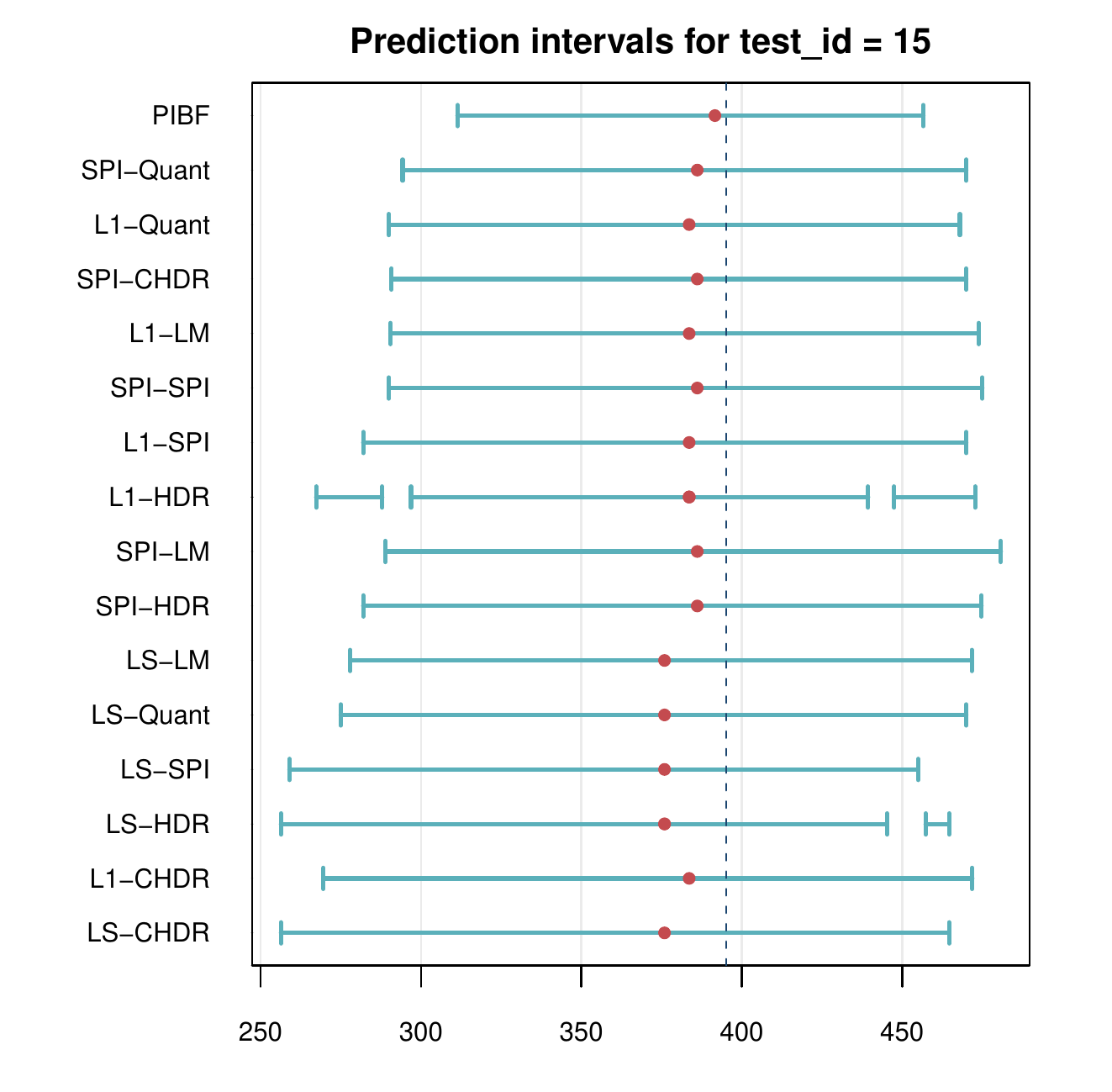}
  \caption{Prediction intervals for the 15th test observation in the test data. The \textit{x}-axis represents the sale price of houses in thousands. For each method, red dots represent the point prediction and blue lines show the prediction interval(s). The vertical dashed line shows the true response value for the test observation. PIBF: Prediction intervals with boosted forests (the proposed method). The notation for the other 15 methods is \textit{split rule}-\textit{PI method}. Splitting rules are LS: Least-squares, L1: $L_1$, SPI: Shortest PI split rule. PI methods are LM: Classical method, Quant: Quantiles, SPI: Shortest PI, HDR: Highest density region, CHDR: Contiguous HDR.}
  \label{figure:piplot}
\end{figure}

Figure~\ref{figure:piplot} presents the prediction intervals and point predictions for the test observation. The methods are ordered in the \textit{y}-axis based on their resulting PI length. For each method, the red point presents the point prediction and blue lines show the constructed prediction interval(s) for the test observation. If the true response of the test observation is known, it is demonstrated with a dashed vertical line. Note that we may have multiple prediction intervals with the HDR PI method. As we can see from the figure, we may have four different point predictions for the same test observation. The PIBF method and the three splitting rules LS, $L_1$ and SPI can produce different point predictions. But all PI method variations for the same splitting rule have the same point prediction. 

\section{Simulation study}
In this section, we compare the predictive performance of the prediction intervals constructed with our proposed method to the existing methods presented in the Introduction using a variety of simulated and real benchmark data sets. 

\subsection{Simulation design}
We apply a simulation study based on seven simulated data sets from the literature. The first three of the data sets are Friedman's benchmark regression problems described in \citet{jerome_h_friedman_multivariate_1991} and \citet{breiman_bagging_1996}. We use the CRAN package \texttt{mlbench} \citep{R-mlbench} to generate these data sets. 

In Friedman Problem 1, the inputs are 10 independent variables uniformly distributed on the interval $\left[0,1\right]$. The first five covariates are used to generate the response:
$$y = 10 \sin\left(\pi x_1 x_2\right) + 20 \left(x_3 - 0.5\right)^2 + 10 x_4 + 5 x_5 + \epsilon$$
where $\epsilon$ is $N\left(0,\sigma^2\right)$ and the default standard deviation of $\epsilon$ is 1 which yields a signal-to-noise ratio (SNR) (\textit{i.e.}, the ratio of the standard deviation of signal to the standard deviation of error) of 4.8:1.

In Friedman Problem 2, the response is generated as $$y = \left(x_1^2 + \left(x_2 x_3 - \frac{1}{x_2 x_4}\right)^2\right)^{0.5} + \epsilon$$ where the inputs are four independent variables uniformly distributed over the ranges 
\begin{align*}
    0 \leq &x_1 \leq 100 \\
    40 \pi \leq &x_2 \leq 560 \pi \\
    0 \leq &x_3 \leq 1\\
    1 \leq &x_4 \leq 11
\end{align*}
and $\epsilon$ is $N\left(0,\sigma^2\right)$. The default value of 125, which yields a SNR of 3:1, is used for the standard deviation of $\epsilon$.

In Friedman Problem 3, the inputs are four independent variables uniformly distributed over the same ranges as Friedman Problem 2. The response is generated as $$y = \arctan{\left(\frac{x_2 x_3 - \frac{1}{x_2 x_4}}{x_1}\right)} + \epsilon$$ where $\epsilon$ is $N\left(0,\sigma^2\right)$ and the default value of 0.01 for the standard deviation of $\epsilon$ is used, which yields a SNR of 3:1.

The fourth data set is the Peak Benchmark Problem which is also from the \texttt{mlbench} package. Let $r=3u$ where $u$ is uniform on $\left[0,1\right]$ and let $x$ be uniformly distributed on the \textit{d}-dimensional sphere of radius $r$. The response is $y=25\exp\left(-0.5r^2\right)$. The default value of $d=20$ dimensions is used.

The fifth one is a modification of Friedman Problem 1, which was used in \citet{hothorn_predictive_2021} in their H2c setup. This data set was designed to have heteroscedasticity. The inputs are 10 independent variables uniformly distributed on the interval $\left[0,1\right]$. The first five covariates are used in the mean function and the unscaled mean function is defined as $$\mu = 10 \sin\left(\pi x_1 x_2\right) + 20 \left(x_3 - 0.5\right)^2 + 10 x_4 + 5 x_5$$ Then, the scaled mean function on the interval $\left[-1.5,1.5\right]$ is $$\mu^{S} = \frac{3\left(\mu - \mu_{min}\right)}{\mu_{max} - \mu_{min}}-1.5 $$ where $\mu_{min}$ and $\mu_{max}$ are the minimum and maximum values of $\mu$ over the sample. The last five covariates are used in the standard deviation function and the unscaled standard deviation function is $$\sigma = 10 \sin\left(\pi x_6 x_7\right) + 20 \left(x_8 - 0.5\right)^2 + 10 x_9 + 5 x_{10}$$ The standard deviation is scaled as $$\sigma^{S}=\exp\left(\frac{3\left(\sigma-\sigma_{min}\right)}{\sigma_{max}-\sigma_{min}}-1.5\right)$$ where $\sigma_{min}$ and $\sigma_{max}$ are the minimum and maximum values of $\sigma$ over the sample. The response is generated as a normal random variable with mean $\mu^{S}$ and standard deviation $\sigma^{S}$.

The last two data sets, which were used in \citet{roy_prediction_2020}, have a tree-based response variable. The inputs are seven independent variables generated from the standard normal distribution. The response is generated with the seven covariates according to a tree model with a depth of three, with eight terminal nodes:
\begin{align*}
    y = & u_1 I\left(x_1<0, x_2<0, x_4<0\right)\\
        & + u_2 I\left(x_1<0, x_2<0, x_4 \geq 0\right)\\
        & + u_3 I\left(x_1<0, x_2 \geq 0, x_5<0\right)\\
        & + u_4 I\left(x_1<0, x_2 \geq 0, x_5 \geq 0\right)\\
        & + u_5 I\left(x_1 \geq 0, x_3<0, x_6<0\right)\\
        & + u_6 I\left(x_1 \geq 0, x_3<0, x_6 \geq 0\right)\\
        & + u_7 I\left(x_1 \geq 0, x_3 \geq 0, x_7<0\right)\\
        & + u_8 I\left(x_1 \geq 0, x_3 \geq 0, x_7 \geq 0\right) + \epsilon
\end{align*}
where the terminal node means are $u=\left(5, 10, 15, 20, 25, 30, 35, 40\right)$ and $I$ is the indicator function. The difference in the two data sets is the distribution of the error. In the first one, $\epsilon$ is generated from a standard normal distribution and in the other it is from an exponential distribution with mean 1. The signal-to-noise ratio is 11.5:1 for both data sets.

We use training sample sizes of $n_{train}=\left \{200,500,1000,5000\right \}$, resulting in 28 scenarios. Each scenario is repeated 500 times. In each run, we generate an independent test set of new observations with $n_{test}= 1000$.

\subsection{Competing methods}

We compare our proposed prediction interval estimator with 10 competing methods which were presented in the Introduction. The first is the $\widehat{PI}_\alpha$ method. We fit the random forest with the \texttt{ranger} package and use the \texttt{forestError} package to build PIs. The second is the OOB method. The \texttt{rfinterval} package is used. The third is the split conformal method. The \texttt{conformalInference} package is used. The fourth is the QRF method and the \texttt{quantregForest} package is used. The fifth is the GRF method. The function \texttt{quantile\_forest} in the \texttt{grf} package is used. 

The last five are variations of \citet{roy_prediction_2020}. To compare the performance of the method variations, a comprehensive simulation study and real data analyses were performed in \cite{roy_prediction_2020}. One of the biggest conclusions from these comparison studies was that, among the three alternative splitting criteria within the CART paradigm, the impact of the choice of the splitting rule on the performance of the prediction intervals was moderate whereas the selection of the PI method had a much greater impact on the performance. Hence, in this simulation study, we set up the splitting rule to the least-squares (LS) and only compare the five PI methods: LM, Quant, SPI, HDR, and CHDR. For those methods, the \texttt{rfpi()} function of the \texttt{RFpredInterval} package is used. We fit the random forest with the \texttt{ranger} package. 

\subsection{Parameter settings}

For the simulations, we use the following parameters. For all methods, we set the number of trees to 2000. Letting $p$ be the number of covariates, then the number of covariates to randomly split at each node, \textit{mtry}, is set to $\max\left \{\lfloor p/3 \rfloor,1\right \}$ (except for the GRF method). For the GRF method, following \citet{athey_generalized_2019}, $mtry$ is set to $\min\left \{\lceil \sqrt{p}+20 \rceil,p\right \}$. Also, we use the honest splitting regime with the default fraction of 0.5 for the GRF method. The minimum node size parameter for all forests is set to 5. The desired coverage is set to $95\%$ for all methods. For the proposed method, we perform the cross-validation-based calibration as the primary calibration procedure, but we also investigate the OOB calibration. For the method variations in \citet{roy_prediction_2020}, we perform the OOB calibration procedure as they proposed. For both calibration procedures, the acceptable range of coverage is set to $\left[.945,.955\right]$. Calibration is not performed for the competing methods since no option for calibration is offered in their CRAN packages.

\subsection{Performance with the simulated data sets}

We can evaluate the performance of the competing methods with two measures: the mean coverage and the mean prediction interval length. Table~\ref{table:simulations} presents the average coverage rate of each method on the test set over 500 replications for a given simulated data set and sample size, with average mean prediction interval lengths shown in parentheses. The principal goal of any prediction interval method is to ensure the desired coverage level. In this simulation study, the desired coverage level is set to $95\%$ for all methods. The left plot in Figure~\ref{figure:global} shows the mean coverages over the 28 scenarios for all methods. Overall, all of the methods, except the QRF and GRF methods which tend to be conservative, provide a mean coverage close to the desired level. QRF and GRF methods have an average mean coverage of 0.975 and 0.974 over all scenarios, respectively. Although the $\widehat{PI}_\alpha$ method has an average of the mean coverages of 0.957, close to the desired level, its variability is large. Over 308 (11 methods $\times$ 28 scenarios) average coverage values, there is only one case where the mean coverage is below 0.94. It corresponds to the $\widehat{PI}_\alpha$ method in Friedman Problem 2 with $n_{train}=5000$.

Once the prediction intervals provide the desired coverage level, the next goal of any PI method is to provide the shortest PI length. Prior to carrying out a detailed comparison of interval lengths, we can globally compare the interval lengths over all scenarios with the percentage increase in mean PI length of a method with respect to the best method for a given run. For a given run, define $ml_{i}$ as the mean PI length of method $i$ and $ml^{*}$ as the shortest mean PI length over the 10 competing methods. The percentage increase in PI length for method $i$ is computed as $100 \times \left(ml_i - ml^{*}\right)/ml^{*}$. Smaller values for this measure indicate better performances. The right plot in Figure~\ref{figure:global} presents the relative lengths of the methods across 14,000 runs (28 scenarios $\times$ 500 replications). The prediction intervals with the GRF method are the widest, followed by the QRF method. However, as we saw in the left plot in Figure~\ref{figure:global}, GRF and QRF produce conservative prediction intervals, so their PI lengths cannot be fairly compared to the other methods with a coverage closer to 0.95. Based on the global results, the proposed method, PIBF, performs the best. Following PIBF, $\widehat{PI}_\alpha$, OOB, LM, SPI, HDR and CHDR perform similarly well, with $\widehat{PI}_\alpha$ being slightly better. Among the variations of \citet{roy_prediction_2020}, the PIs with quantiles produce longer prediction intervals than the other four variations. 

\begin{widefigure}[htbp]
  \centering
  \includegraphics[scale=0.7]{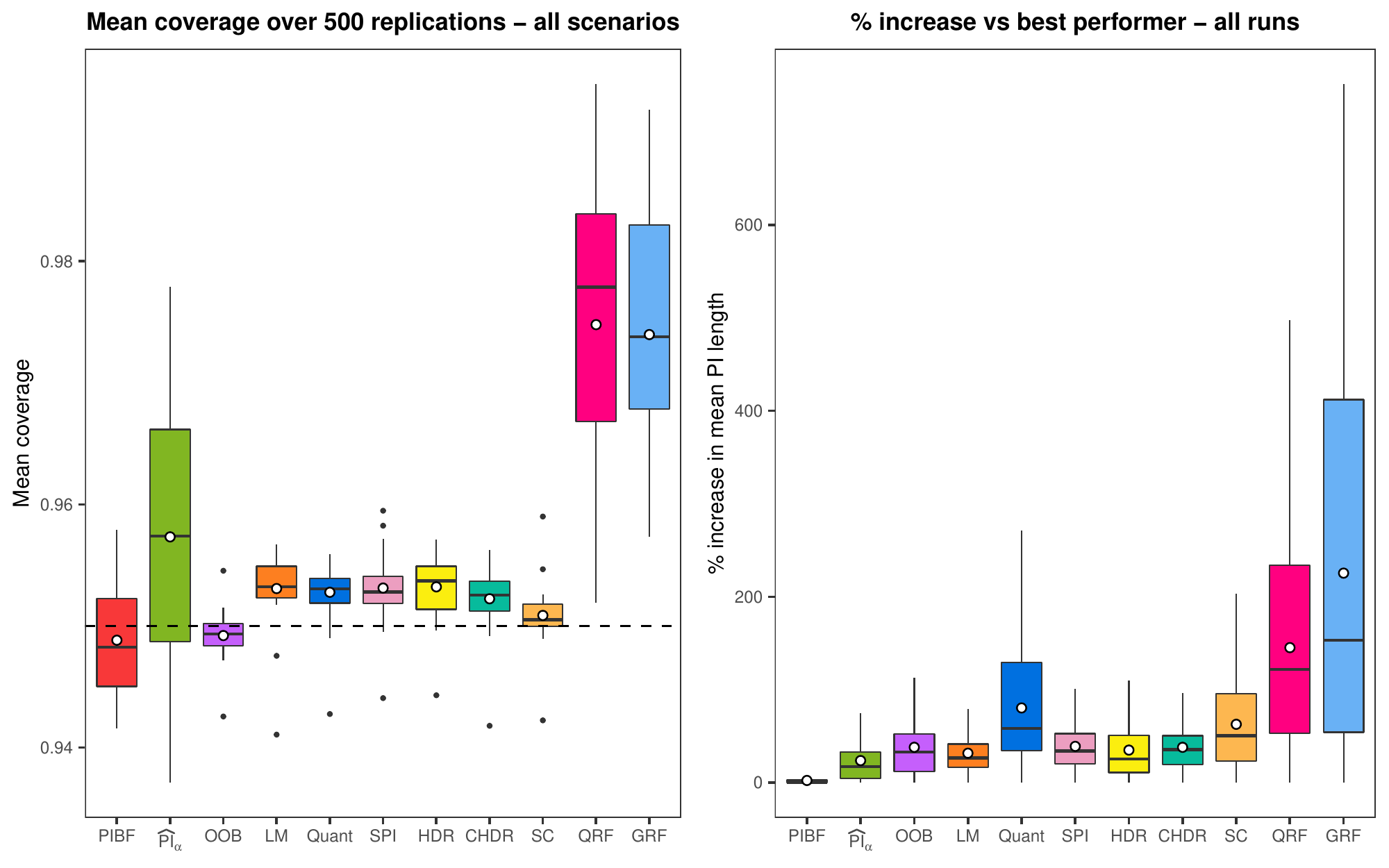}
  \caption{\textit{(Left)} Boxplots for the mean coverage over all scenarios. All methods, except the QRF and GRF methods, are able to provide a mean coverage close to the desired coverage level of 0.95. Each white circle is the average of the mean coverages over 28 scenarios. \textit{(Right)} Boxplots for the percentage increase in mean PI length of each method compared to the shortest PI length for a given run across 14,000 runs. The smallest is the percentage increase, the better is the method. Each white circle is the average of the relative lengths over 14,000 runs. Since the outlier values are distorting the scales, they are removed from the graph. PIBF: Prediction intervals with boosted forests (the proposed method), $\widehat{PI}_\alpha$: Conditional $\alpha$-level prediction interval, OOB: Out-of-Bag approach, LM: Classical method, Quant: Quantiles, SPI: Shortest PI, HDR: Highest density region, CHDR: Contiguous HDR, SC: Split conformal, QRF: Quantile regression forest, GRF: Generalized random forest.}
  \label{figure:global}
\end{widefigure}

Now, we investigate the performance of the methods separately for each scenario. See figures~\ref{figure:dgp1} to \ref{figure:dgp7} in the Supplementary Material for the mean PI length results of each method for each simulated data set. Each figure has four facets corresponding to the four levels of the training sample size. For all methods and data sets, the mean PI lengths and their variability decrease as the sample size increases (except the GRF method for the tree-based data sets). We see that for Friedman Problem 1, from Figure~\ref{figure:dgp1}, for all sample sizes, PIBF consistently outperforms the 10 competing methods in terms of mean PI length while ensuring the desired coverage level (see Table~\ref{table:simulations} for the mean coverage results). QRF and GRF provide the widest prediction intervals for all sample sizes. However, as presented in Table~\ref{table:simulations}, these methods heavily over-cover and are therefore not comparable with the other methods. While the $\widehat{PI}_\alpha$ method also slightly over-covers, it has shorter PIs than other methods.

For Friedman Problem 2 (see Figure~\ref{figure:dgp2} in the Supplementary Material), we see that the proposed method has the shortest mean PI length for the smallest sample size, and as the sample size increases $\widehat{PI}_\alpha$ provides shorter PIs. However, we should also take into account the mean coverages presented in Table~\ref{table:simulations}. The proposed method has smaller coverage levels for $n_{train}=200$ compared to $\widehat{PI}_\alpha$, but as the sample size increases the coverage levels decrease for the $\widehat{PI}_\alpha$ method (up to 0.937 for $n_{train}=5000$) whereas PIBF keeps it around 0.945. For $n_{train}=5000$, the OOB method builds shorter PIs while ensuring the desired coverage level. Again, QRF and GRF have the widest PIs for all sample sizes due to their conservative PIs.

The performance of PIBF and $\widehat{PI}_\alpha$ is very similar for Friedman Problem 3 (see Figure~\ref{figure:dgp3} in the Supplementary Material). Both methods provide the shortest PIs with similar coverage levels and mean PI lengths. Results for the Peak Benchmark Problem presented in Figure~\ref{figure:dgp4} are very similar to those of Friedman Problem 1. For all sample sizes, PIBF consistently outperforms the 10 competing methods in terms of mean PI length. But this time, SPI method with LS splitting rule comes in second place.

For the H2c setup (see Figure~\ref{figure:dgp5} in the Supplementary Material), which is the modification of Friedman Problem 1, we can see that all methods are comparable since QRF and GRF do not over-cover. In this setting, all methods also perform fairly well with respect to PI length. Overall, the LM prediction interval method with the LS splitting rule provides slightly shorter PIs.

For the tree-based data sets (see figures~\ref{figure:dgp6} and \ref{figure:dgp7} in the Supplementary Material), overall, it seems that the distribution of the error does not have a significant effect on the results. Again, QRF and GRF have conservative PIs. Unlike the other data sets, we see here that the mean PI lengths of the GRF method decrease very slowly as the sample size increases. For $n_{train}=5000$, all methods (except QRF and GRF) perform similarly. For the smallest sample size, HDR PI building methods and the proposed method perform slightly better than other methods. As the sample size increases, PIBF produces the shortest prediction intervals.

\begin{widetable}[t]
\caption{\label{table:simulations}Results of the simulation study. Average coverage rates of each method in each simulation, with average mean PI lengths shown in parentheses. The desired coverage level is 0.95. Shortest average mean PI lengths are emphasized with bold text. PIBF: Prediction intervals with boosted forests (the proposed method), $\widehat{PI}_\alpha$: Conditional $\alpha$-level prediction interval, OOB: Out-of-Bag approach, LM: Classical method, Quant: Quantiles, SPI: Shortest PI, HDR: Highest density region, CHDR: Contiguous HDR, SC: Split conformal, QRF: Quantile regression forest, GRF: Generalized random forest.}
\centering
\resizebox{\linewidth}{!}{
\begin{tabular}[t]{rllllllllllll}
\toprule
$n_{train}$ & Data & PIBF & $\widehat{PI}_\alpha$ & OOB & LM & Quant & SPI & HDR & CHDR & SC & QRF & GRF\\
\midrule
 & Friedman 1 & 0.952 (\textbf{7.69}) & 0.959 (9.7) & 0.949 (10.3) & 0.955 (11.2) & 0.956 (12.5) & 0.956 (12.1) & 0.955 (12.2) & 0.956 (11.5) & 0.952 (12) & 0.978 (15.3) & 0.968 (17.7)\\

 & Friedman 2 & 0.942 (\textbf{635}) & 0.951 (679) & 0.947 (762) & 0.956 (772) & 0.955 (990) & 0.956 (898) & 0.955 (793) & 0.954 (800) & 0.951 (912) & 0.969 (1103) & 0.972 (1241)\\

 & Friedman 3 & 0.944 (\textbf{0.57}) & 0.948 (0.62) & 0.949 (0.75) & 0.956 (0.69) & 0.953 (0.83) & 0.952 (0.74) & 0.951 (0.88) & 0.954 (0.76) & 0.953 (0.93) & 0.961 (0.89) & 0.968 (0.88)\\

 & Peak & 0.958 (\textbf{9.06}) & 0.956 (12.4) & 0.949 (13.3) & 0.955 (12) & 0.956 (17.9) & 0.957 (10.5) & 0.955 (15.4) & 0.955 (11.1) & 0.951 (15.5) & 0.972 (20.6) & 0.978 (20.2)\\

 & H2c & 0.954 (6.5) & 0.954 (6.3) & 0.955 (6.26) & 0.955 (\textbf{5.86}) & 0.956 (6.42) & 0.957 (6.47) & 0.956 (6.61) & 0.955 (6.45) & 0.959 (6.68) & 0.952 (6.09) & 0.957 (6.35)\\

 & Tree-N & 0.949 (9.12) & 0.967 (12.3) & 0.947 (14.2) & 0.956 (11.9) & 0.956 (22.3) & 0.958 (15.4) & 0.955 (\textbf{9.08}) & 0.955 (11.9) & 0.951 (18.9) & 0.979 (29.6) & 0.960 (36)\\

\multirow[t]{-7}{*}{\raggedleft\arraybackslash 200} & Tree-exp & 0.949 (\textbf{9.31}) & 0.967 (12.4) & 0.947 (14.2) & 0.955 (11.9) & 0.956 (22.1) & 0.959 (15.1) & 0.956 (10.4) & 0.956 (12.5) & 0.950 (18.9) & 0.979 (29.5) & 0.960 (35.8)\\
\cmidrule{1-13}
 & Friedman 1 & 0.952 (\textbf{6.32}) & 0.962 (8.25) & 0.948 (8.81) & 0.953 (9.37) & 0.952 (10.1) & 0.953 (9.87) & 0.952 (9.81) & 0.953 (9.47) & 0.951 (10.1) & 0.986 (14) & 0.978 (16.9)\\

 & Friedman 2 & 0.945 (586) & 0.946 (\textbf{585}) & 0.948 (650) & 0.952 (676) & 0.952 (820) & 0.952 (744) & 0.951 (679) & 0.951 (690) & 0.952 (751) & 0.975 (998) & 0.979 (1120)\\

 & Friedman 3 & 0.943 (\textbf{0.5}) & 0.946 (0.53) & 0.948 (0.62) & 0.952 (0.61) & 0.951 (0.71) & 0.951 (0.65) & 0.951 (0.7) & 0.951 (0.67) & 0.952 (0.75) & 0.965 (0.79) & 0.970 (0.77)\\

 & Peak & 0.957 (\textbf{6.44}) & 0.967 (9.97) & 0.951 (11) & 0.953 (9.75) & 0.954 (15) & 0.954 (8.22) & 0.956 (11.6) & 0.953 (8.93) & 0.952 (12.8) & 0.980 (18.7) & 0.984 (17.8)\\

 & H2c & 0.953 (5.81) & 0.956 (5.88) & 0.951 (5.89) & 0.952 (\textbf{5.43}) & 0.953 (5.83) & 0.953 (5.81) & 0.954 (5.92) & 0.952 (5.84) & 0.955 (6.17) & 0.954 (5.79) & 0.959 (5.94)\\

 & Tree-N & 0.948 (\textbf{6.16}) & 0.969 (8.17) & 0.949 (9.85) & 0.952 (8.23) & 0.953 (15.5) & 0.953 (10) & 0.954 (7.31) & 0.952 (9.6) & 0.952 (13.3) & 0.984 (25.8) & 0.969 (35.7)\\

\multirow[t]{-7}{*}{\raggedleft\arraybackslash 500} & Tree-exp & 0.947 (\textbf{6.31}) & 0.966 (8.3) & 0.949 (9.93) & 0.952 (8.19) & 0.953 (15.4) & 0.954 (9.67) & 0.954 (7.93) & 0.953 (9.96) & 0.953 (13.4) & 0.984 (25.8) & 0.970 (35.5)\\
\cmidrule{1-13}
 & Friedman 1 & 0.953 (\textbf{5.67}) & 0.964 (7.42) & 0.949 (7.95) & 0.954 (8.37) & 0.954 (8.89) & 0.954 (8.75) & 0.954 (8.65) & 0.953 (8.4) & 0.950 (8.85) & 0.990 (12.9) & 0.985 (15.9)\\

 & Friedman 2 & 0.945 (565) & 0.942 (\textbf{546}) & 0.950 (594) & 0.953 (635) & 0.953 (735) & 0.953 (682) & 0.952 (637) & 0.953 (645) & 0.951 (658) & 0.977 (910) & 0.983 (1022)\\

 & Friedman 3 & 0.944 (\textbf{0.47}) & 0.945 (0.48) & 0.948 (0.55) & 0.954 (0.58) & 0.952 (0.65) & 0.952 (0.61) & 0.951 (0.63) & 0.952 (0.63) & 0.949 (0.63) & 0.967 (0.73) & 0.971 (0.72)\\

 & Peak & 0.957 (\textbf{5}) & 0.972 (8.44) & 0.950 (9.54) & 0.952 (8.51) & 0.953 (13.2) & 0.953 (7.21) & 0.956 (9.35) & 0.952 (7.82) & 0.951 (11) & 0.983 (17.5) & 0.986 (16.7)\\

 & H2c & 0.952 (5.48) & 0.955 (5.64) & 0.949 (5.69) & 0.948 (\textbf{5.17}) & 0.949 (5.49) & 0.950 (5.51) & 0.950 (5.49) & 0.949 (5.49) & 0.949 (5.78) & 0.954 (5.61) & 0.958 (5.71)\\

 & Tree-N & 0.946 (\textbf{5.11}) & 0.965 (6.2) & 0.950 (7.52) & 0.954 (6.67) & 0.953 (12.1) & 0.954 (7.91) & 0.954 (6.21) & 0.953 (8.33) & 0.950 (9.94) & 0.985 (21.6) & 0.976 (35.1)\\

\multirow[t]{-7}{*}{\raggedleft\arraybackslash 1000} & Tree-exp & 0.946 (\textbf{5.16}) & 0.964 (6.39) & 0.949 (7.64) & 0.954 (6.62) & 0.953 (11.9) & 0.954 (7.45) & 0.955 (6.5) & 0.953 (8.45) & 0.950 (10) & 0.985 (21.6) & 0.975 (34.9)\\
\cmidrule{1-13}
 & Friedman 1 & 0.950 (\textbf{4.71}) & 0.967 (6.04) & 0.950 (6.47) & 0.953 (6.67) & 0.953 (6.97) & 0.953 (6.91) & 0.954 (6.78) & 0.953 (6.67) & 0.950 (7.03) & 0.995 (10.8) & 0.992 (13.2)\\

 & Friedman 2 & 0.945 (543) & 0.937 (\textbf{507}) & 0.950 (529) & 0.952 (574) & 0.951 (610) & 0.951 (592) & 0.951 (570) & 0.951 (573) & 0.950 (549) & 0.975 (717) & 0.982 (781)\\

 & Friedman 3 & 0.945 (0.44) & 0.941 (\textbf{0.43}) & 0.950 (0.46) & 0.953 (0.53) & 0.952 (0.56) & 0.952 (0.55) & 0.951 (0.54) & 0.952 (0.56) & 0.950 (0.49) & 0.966 (0.62) & 0.971 (0.62)\\

 & Peak & 0.955 (\textbf{2.84}) & 0.978 (5.92) & 0.952 (7.11) & 0.952 (6.54) & 0.954 (10.1) & 0.952 (5.68) & 0.957 (6.46) & 0.951 (6.04) & 0.950 (8) & 0.988 (14.9) & 0.990 (15.2)\\

 & H2c & 0.949 (5) & 0.953 (5.31) & 0.943 (5.39) & 0.941 (\textbf{4.82}) & 0.943 (5.01) & 0.944 (5.06) & 0.944 (5.03) & 0.942 (4.98) & 0.942 (5.41) & 0.953 (5.33) & 0.958 (5.31)\\

 & Tree-N & 0.945 (\textbf{4.33}) & 0.949 (4.37) & 0.951 (4.77) & 0.957 (5.25) & 0.951 (7.03) & 0.950 (5.4) & 0.953 (4.78) & 0.951 (5.89) & 0.950 (5.55) & 0.979 (10.5) & 0.986 (32.1)\\

\multirow[t]{-7}{*}{\raggedleft\arraybackslash 5000} & Tree-exp & 0.945 (\textbf{4.08}) & 0.959 (4.39) & 0.950 (4.91) & 0.955 (5.12) & 0.951 (6.87) & 0.952 (4.78) & 0.954 (4.36) & 0.949 (5.49) & 0.950 (5.73) & 0.978 (10.5) & 0.986 (31.8)\\
\bottomrule
\end{tabular}}
\end{widetable}

\subsection{Effect of calibration on the performance of prediction intervals}

\begin{figure}[htbp]
  \centering
  \includegraphics[scale=0.8]{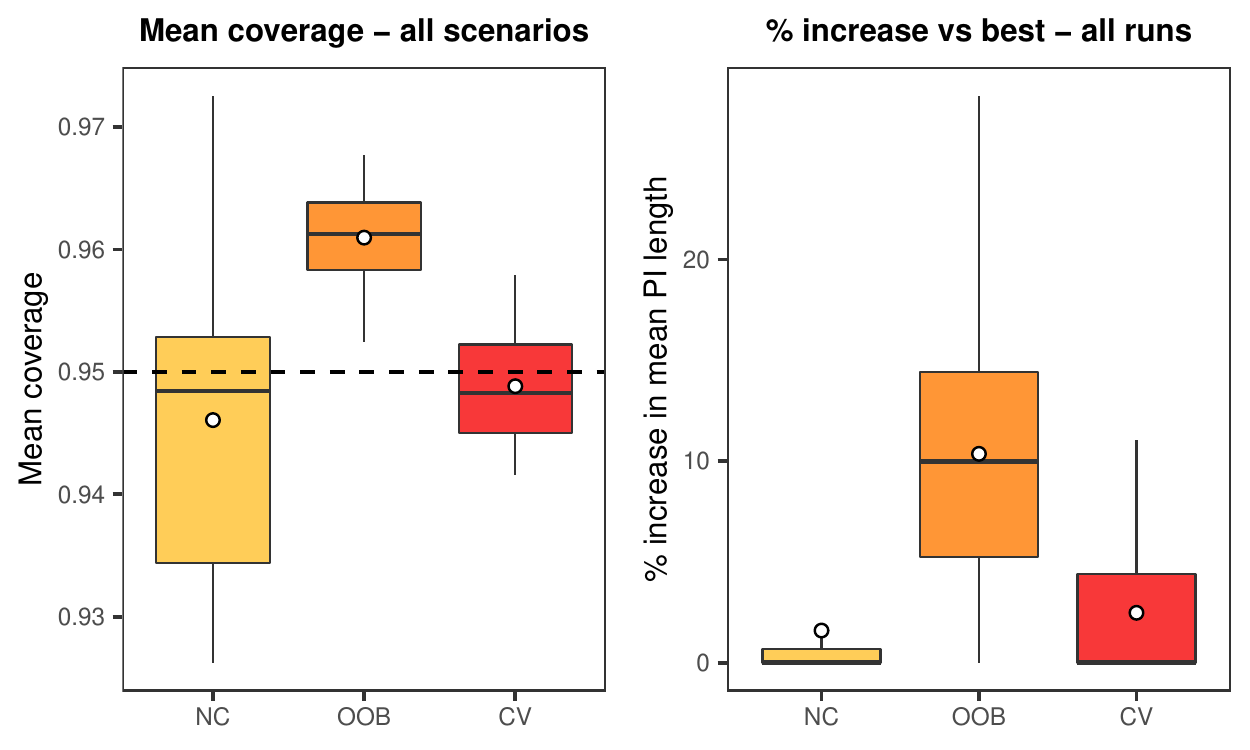}
  \caption{Global performance results of the proposed method for the simulated data sets with different calibration procedures. NC: No calibration, OOB: OOB calibration, CV: Calibration with cross-validation. \textit{(Left)} Boxplots for the mean coverage over 500 replications across all scenarios. \textit{(Right)} Boxplots for the percentage increase in mean PI length of each calibration procedure compared to the shortest PI length for a given run across 14,000 runs. Smaller values are better. Since outlier values are distorting the scales, they are removed. }
  \label{figure:calib}
\end{figure}

In this section, we investigate the effect of the proposed calibration on performance of the prediction intervals. We apply the same simulation study using the seven simulated data sets. We compare the results of the proposed method without calibration, with OOB calibration and calibration with cross-validation. The desired coverage level is set to 95\%. In Figure~\ref{figure:calib}, the left plot presents the mean coverages over 28 scenarios for the three variations, and the right plot shows the percentage increase in mean PI length of each of the three calibration variants across 14,000 runs (28 scenarios $\times$ 500 replications). Although we obtain the shortest prediction intervals without calibration, the variability of the mean coverage level is larger and sometimes the coverage falls below 0.94.  Looking at the left plot, we can say that the variability of the mean coverage level decreases with both calibration procedures. However, applying OOB calibration provides conservative PIs. The median of the mean coverage level is more than 0.96 and the PIs with the OOB calibration are the widest. Applying calibration with CV produces slightly longer PIs than those with no calibration, but these PIs have coverage levels closer to the desired level. 

In the package, both calibration procedures are implemented for the PIBF method. Simulation study results show that, compared to the OOB calibration, calibration with CV produces shorter PIs while maintaining the desired coverage level. Therefore, the default calibration procedure is set to CV in the \texttt{pibf()} function. In terms of computational time (see tables~\ref{table:timesim} and \ref{table:timereal}), calibration with $k$-fold CV is slower than OOB calibration since it needs to fit two additional random forests for each fold. 

\subsection{Performance with real data sets}

To further explore the performance of the prediction intervals built with the proposed method, we use 11 real data sets. Since two of the data sets have two response variables, we consider that we are analyzing 13 real data sets. Boston housing and Ames housing data sets are from the R packages \texttt{mlbench} and \texttt{AmesHousing}, respectively. The other data sets are obtained through the UCI Machine Learning Repository \citep{ucirepo}.

\begin{widefigure}[htbp]
  \centering
  \includegraphics[scale=0.7]{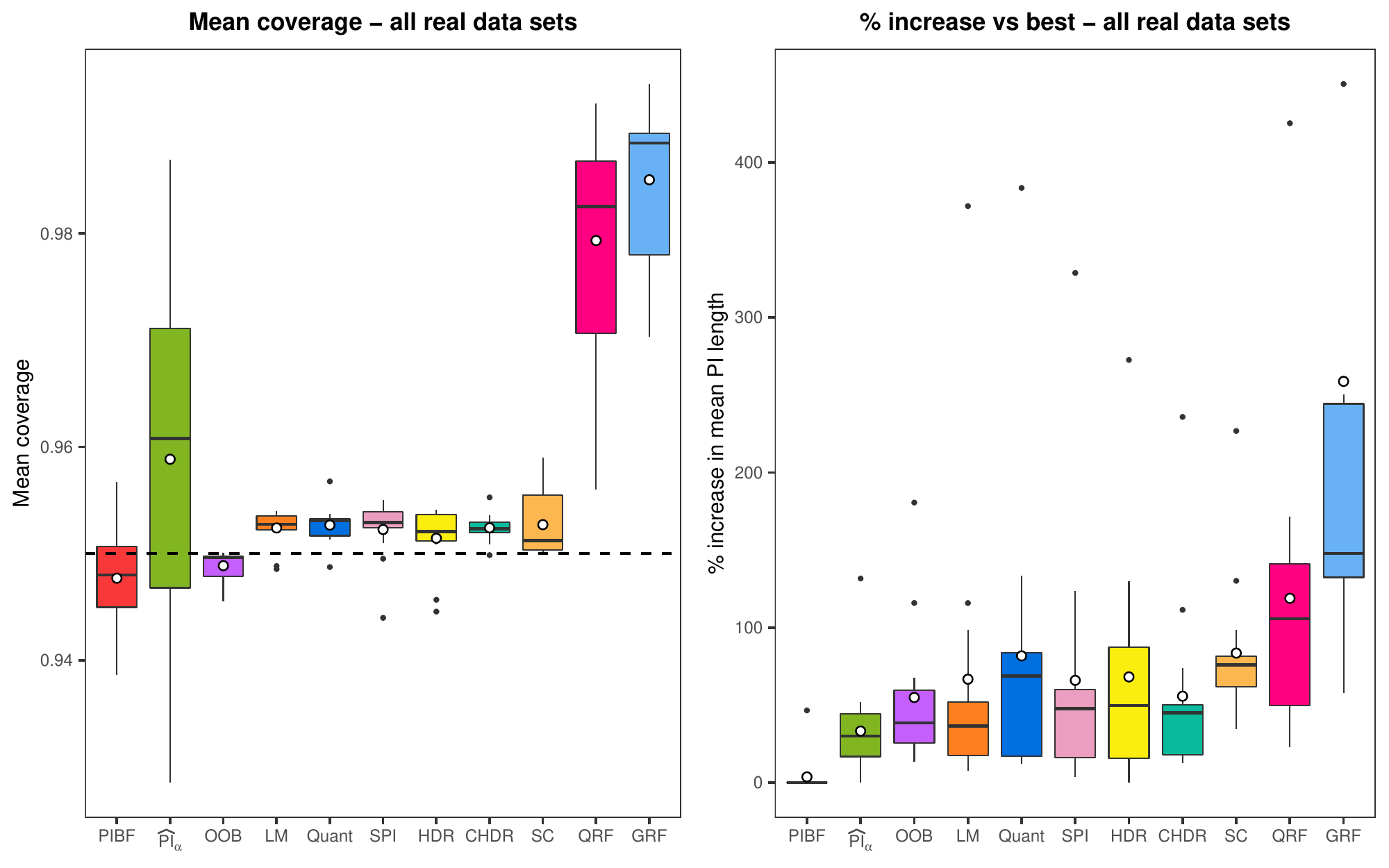}
  \caption{\textit{(Left)} Boxplots for the mean coverage over all real data sets. All methods except the QRF and GRF methods are able to provide a mean coverage close to the desired coverage level of 0.95. Each white circle is the average of the mean coverages over 13 real data sets. \textit{(Right)} Boxplots for the percentage increase in mean PI length of each method compared to the shortest PI length for a given real data set across 13 data sets. The smallest the percentage increase, the better the method. Each white circle is the average of the relative lengths over 13 real data sets. One of the outlier values for GRF with the percentage increase of 1244\% is removed from the graph since it is distorting the scales. PIBF: Prediction intervals with boosted forests (the proposed method), $\widehat{PI}_\alpha$: Conditional $\alpha$-level prediction interval, OOB: Out-of-Bag approach, LM: Classical method, Quant: Quantiles, SPI: Shortest PI, HDR: Highest density region, CHDR: Contiguous HDR, SC: Split conformal, QRF: Quantile regression forest, GRF: Generalized random forest.}
  \label{figure:globalreal}
\end{widefigure}

For each data set, we apply 100 times 10-fold cross-validation for each method. Hence, for each fold, the training and testing sets correspond to 90\% and 10\% of the whole data set, respectively. The desired coverage level is set to 95\% for all methods. Table~\ref{table:realdata} presents the results of the real data analyses (\textit{n} is the number of observations and \textit{p} is the number of predictors) with the average coverage rate of each method over 100 repetitions, and mean prediction interval lengths averaged over 100 repetitions shown in parentheses. Figure~\ref{figure:globalreal} illustrates the global results of the analyses across datasets. The left plot in Figure~\ref{figure:globalreal} shows the mean coverages over the 13 real data sets for all methods. Similar to what we have seen with the simulated data sets, the QRF and GRF methods produce conservative prediction intervals, whereas the other methods provide a mean coverage close to the desired level. Again, the $\widehat{PI}_\alpha$ method maintains the target level on average with 0.959, but its variability is the highest among all methods. Across all data sets, there are three cases where the mean coverage is below 0.94: the proposed method for Concrete slump, and the $\widehat{PI}_\alpha$ method for Auto MPG and Computer hardware. 

The right plot in Figure~\ref{figure:globalreal} presents the relative lengths of methods across 13 real data sets. For each method, there are 13 points in the boxplot, and each point corresponds to the percentage increase in mean PI length compared to the best method for a single real data set. Again, the prediction intervals with GRF and QRF are the widest among eleven methods. Among the other nine methods, the proposed method performs the best, followed by $\widehat{PI}_\alpha$. 

For each real data set, we analyze the performance of each method through the mean PI lengths presented in figures~\ref{figure:realdata1} to \ref{figure:realdata3} in the Supplementary Material. For the Abalone data set, the HDR method produces the shortest prediction intervals, followed by the SPI and Quant methods. While the QRF method over-covers, its PIs are no wider than those of most of the other methods. The proposed method, PIBF, is distinctly the best prediction interval method yielding the shortest PI lengths for the Air quality data set with absolute and relative humidity response variables, Airfoil selfnoise, Ames housing, Boston housing, Concrete compression, Energy efficiency data set with cooling and heating load response variables, and Servo data sets. In the Auto MPG data set, $\widehat{PI}_\alpha$ has the shortest mean PI length but with a mean coverage of 0.929. Among the other methods, PIBF, OOB, LM, Quant and SPI methods show similarly good performances while maintaining the desired coverage level. For the Computer hardware data set, PIBF, $\widehat{PI}_\alpha$, and SPI methods perform better than the other methods. They have similar mean PI lengths. For the Concrete slump data set, the proposed method has the shortest mean PI length, but with a slightly smaller coverage of 0.939. This data set is the only one of the simulated and real data sets where the proposed method has a mean coverage below 0.94. After PIBF, $\widehat{PI}_\alpha$ and LM show a good performance with a mean coverage close to the target level. Overall, we can conclude that the proposed method shows better performance than the 10 competing methods for almost all of the real data sets.

\begin{widetable}[t]

\caption{\label{table:realdata}Results of the real data analysis. Average coverage rates of each method for each real data set, with average mean PI lengths shown in parentheses. The desired coverage level is 0.95. Shortest average mean PI lengths are shown in bold. PIBF: Prediction intervals with boosted forests (the proposed method), $\widehat{PI}_\alpha$: Conditional $\alpha$-level prediction interval, OOB: Out-of-Bag approach, LM: Classical method, Quant: Quantiles, SPI: Shortest PI, HDR: Highest density region, CHDR: Contiguous HDR, SC: Split conformal, QRF: Quantile regression forest, GRF: Generalized random forest.}
\centering
\resizebox{\linewidth}{!}{
\begin{tabular}[t]{lrrlllllllllll}
\toprule
Data & \textit{n} & \textit{p} & PIBF & $\widehat{PI}_\alpha$ & OOB & LM & Quant & SPI & HDR & CHDR & SC & QRF & GRF\\
\midrule
Abalone & 4177 & 8 & 0.947 (8.18) & 0.946 (8.05) & 0.950 (8.9) & 0.949 (8.09) & 0.953 (6.81) & 0.951 (6.73) & 0.946 (\textbf{5.58}) & 0.951 (8.12) & 0.950 (9.04) & 0.971 (8.03) & 0.978 (8.82)\\

Air quality (AH) &  &  & 0.948 (\textbf{0.22}) & 0.966 (0.29) & 0.950 (0.34) & 0.954 (0.29) & 0.954 (0.33) & 0.954 (0.31) & 0.954 (0.34) & 0.954 (0.29) & 0.950 (0.4) & 0.992 (0.55) & 0.994 (0.78)\\

Air quality (RH) & \multirow{-2}{*}{\raggedleft\arraybackslash 6941} & \multirow{-2}{*}{\raggedleft\arraybackslash 13} & 0.949 (\textbf{12.9}) & 0.971 (18.7) & 0.950 (21.6) & 0.953 (19.6) & 0.953 (22.5) & 0.954 (20.6) & 0.954 (21.1) & 0.953 (19.3) & 0.950 (25.5) & 0.989 (35) & 0.989 (44.3)\\

Airfoil selfnoise & 1503 & 5 & 0.946 (\textbf{8.47}) & 0.972 (12.9) & 0.950 (13.3) & 0.952 (18.3) & 0.952 (19.8) & 0.953 (18.9) & 0.952 (19.5) & 0.953 (17.9) & 0.951 (14.7) & 0.987 (20.4) & 0.988 (18.1)\\

Ames housing & 2929 & 80 & 0.955 (\textbf{69.3}) & 0.961 (81.6) & 0.950 (90.3) & 0.954 (79.3) & 0.953 (80.5) & 0.953 (80.5) & 0.952 (80.3) & 0.952 (80.4) & 0.951 (96.4) & 0.987 (118) & 0.993 (172)\\

Auto MPG & 392 & 7 & 0.945 (10) & 0.929 (\textbf{9.76}) & 0.947 (11.1) & 0.953 (10.5) & 0.953 (10.9) & 0.953 (10.5) & 0.954 (13.2) & 0.952 (11.5) & 0.955 (13.1) & 0.967 (12) & 0.977 (16.1)\\

Boston housing & 506 & 13 & 0.942 (\textbf{10.5}) & 0.948 (11.2) & 0.949 (12.3) & 0.954 (11.7) & 0.953 (11.8) & 0.953 (11.4) & 0.952 (12.2) & 0.952 (12) & 0.951 (15) & 0.982 (15.7) & 0.990 (25.9)\\

Computer hardware & 209 & 6 & 0.942 (\textbf{139}) & 0.939 (140) & 0.946 (171) & 0.952 (163) & 0.951 (163) & 0.950 (144) & 0.951 (157) & 0.950 (156) & 0.957 (248) & 0.965 (182) & 0.985 (365)\\

Concrete compression & 1030 & 8 & 0.948 (\textbf{14.8}) & 0.957 (18) & 0.950 (19.6) & 0.954 (20.2) & 0.953 (27.2) & 0.954 (22.8) & 0.953 (27.7) & 0.953 (21.5) & 0.950 (26) & 0.982 (34.5) & 0.989 (46.6)\\

Concrete slump & 103 & 7 & 0.939 (\textbf{12.2}) & 0.947 (14.2) & 0.948 (15.3) & 0.949 (14.9) & 0.949 (20.6) & 0.944 (19.4) & 0.945 (16.3) & 0.951 (15.4) & 0.958 (22.1) & 0.956 (22.3) & 0.970 (28.3)\\

Energy efficiency (CL) &  &  & 0.953 (\textbf{3.71}) & 0.977 (5) & 0.950 (8.01) & 0.952 (7.37) & 0.952 (7.9) & 0.952 (7.04) & 0.951 (8.03) & 0.953 (6.46) & 0.951 (8.54) & 0.985 (8.18) & 0.986 (20.4)\\

Energy efficiency (HL) & \multirow{-2}{*}{\raggedleft\arraybackslash 768} & \multirow{-2}{*}{\raggedleft\arraybackslash 8} & 0.951 (\textbf{1.48}) & 0.987 (3.43) & 0.950 (4.16) & 0.953 (6.99) & 0.952 (7.17) & 0.954 (6.35) & 0.951 (5.52) & 0.952 (4.98) & 0.952 (4.84) & 0.988 (7.79) & 0.989 (19.9)\\

Servo & 167 & 4 & 0.957 (\textbf{18.4}) & 0.966 (24.2) & 0.948 (25.5) & 0.953 (26.6) & 0.957 (32.9) & 0.955 (27.2) & 0.954 (28.2) & 0.955 (26.8) & 0.959 (30.4) & 0.983 (37.9) & 0.977 (42.9)\\
\bottomrule
\end{tabular}}
\end{widetable}

\subsection{Comparison of the computational times}

All simulations and real data analyses were conducted in R version 3.6.0 on a Linux machine with Intel(R) Xeon(R) E5-2667 v3 @ 3.20GHz with 396 GB of memory. The average computational time of each method for the simulated and real data sets are presented in tables~\ref{table:timesim} and \ref{table:timereal}. For the proposed method (PIBF), the computational times for both calibration methods, cross-validation and OOB, are presented in the tables. We can see from the tables that, for most of the data sets, calibration with cross-validation has longer running times than OOB calibration, which is expected since with the \textit{k}-fold cross-validation, we fit $2k$ more random forests than applying OOB calibration.

For the variations of \cite{roy_prediction_2020}, since we can build prediction intervals with the five PI methods by only fitting a single random forest  with a selected splitting rule, we present the total computational time for building the five variations under RFPI. To be clear, for a given splitting rule, the \texttt{rfpi()} function fits a random forest and then the set of PI methods requested by the user are applied to the output of the random forest. In our simulations, we choose to return all five PI methods for the selected splitting rule, \textit{i.e.} when we measure the running time of the \texttt{rfpi()} function, we get the total running time of building five prediction intervals. Therefore, we should interpret the values for RFPI with care while comparing the computational times of the methods. Although not all PI methods have similar computational complexities, we can say that even the average time of building prediction intervals with one of these variations, assuming they have similar running times, is reasonable. Since, for the HDR-based PI methods, an optimal bandwidth has to be chosen, which is a time-consuming process, among the five PI methods, the slowest ones are the HDR and CHDR. From the remaining variations, the classical method, LM, is the fastest, followed by the Quant and SPI methods.

For both the simulations and real data analyses, the OOB and GRF methods have the smallest running times. For most of the methods, the increase in the sample size has a mild effect on the ratio of increase in running times. However, for the split conformal method with the simulated data sets, running times increase more than the proportional increase in sample sizes. Similarly, we can see that the QRF method is also affected from the training sample size as it rises to 5000.

\begin{table}[t]

\caption{\label{table:timesim}Average computational time (in seconds) of each method over 500 replications for each simulated data set. The values represent the average time to build prediction intervals for a test set with 1000 observations. PIBF-CV: The proposed method with the cross-validation calibration, PIBF-OOB: The proposed method with the OOB calibration, RFPI: The average total running time of the five PI methods, \textit{i.e.} LM + Quant + SPI + HDR + CHDR.}
\centering
\resizebox{0.81\linewidth}{!}{
\begin{tabular}[t]{rlrrrrrrrr}
\toprule
$n_{train}$ & Data & PIBF-CV & PIBF-OOB & $\widehat{PI}_\alpha$ & OOB & RFPI & SC & QRF & GRF\\
\midrule
 & Friedman 1 & 21.36 & 13.44 & 9.62 & 0.25 & 87.79 & 0.61 & 3.05 & 0.27\\

 & Friedman 2 & 21.83 & 13.96 & 9.62 & 0.23 & 59.50 & 0.44 & 2.61 & 0.26\\

 & Friedman 3 & 28.84 & 16.03 & 11.46 & 0.20 & 75.47 & 0.41 & 2.50 & 0.22\\

 & Peak & 18.76 & 10.95 & 11.91 & 0.28 & 73.15 & 0.89 & 4.05 & 0.43\\

 & H2c & 12.92 & 9.83 & 7.99 & 0.27 & 36.94 & 0.79 & 4.05 & 0.30\\

 & Tree-N & 22.75 & 21.17 & 12.21 & 0.23 & 100.02 & 0.50 & 2.81 & 0.26\\

\multirow[t]{-7}{*}{\raggedleft\arraybackslash 200} & Tree-exp & 16.28 & 15.62 & 15.65 & 0.23 & 87.49 & 0.53 & 2.85 & 0.25\\
\cmidrule{1-10}
 & Friedman 1 & 38.37 & 14.63 & 8.50 & 0.44 & 96.18 & 2.45 & 9.29 & 0.66\\

 & Friedman 2 & 27.07 & 13.64 & 7.64 & 0.43 & 84.19 & 1.80 & 6.79 & 0.47\\

 & Friedman 3 & 18.85 & 17.49 & 7.98 & 0.47 & 70.78 & 1.86 & 4.96 & 0.45\\

 & Peak & 29.16 & 18.14 & 9.72 & 0.70 & 212.23 & 2.29 & 7.49 & 1.26\\

 & H2c & 14.84 & 11.29 & 11.20 & 0.44 & 58.60 & 1.74 & 4.95 & 0.78\\

 & Tree-N & 17.80 & 20.31 & 7.70 & 0.51 & 183.92 & 2.21 & 4.28 & 0.51\\

\multirow[t]{-7}{*}{\raggedleft\arraybackslash 500} & Tree-exp & 18.05 & 19.05 & 7.74 & 0.51 & 170.24 & 1.72 & 4.45 & 0.49\\
\cmidrule{1-10}
 & Friedman 1 & 32.92 & 18.07 & 12.11 & 0.64 & 181.30 & 3.35 & 9.54 & 1.50\\

 & Friedman 2 & 23.07 & 15.94 & 10.25 & 0.45 & 135.64 & 2.14 & 6.26 & 0.90\\

 & Friedman 3 & 23.24 & 16.07 & 7.37 & 0.44 & 96.76 & 2.27 & 6.40 & 0.69\\

 & Peak & 51.93 & 33.12 & 7.95 & 0.83 & 374.89 & 5.35 & 19.71 & 1.65\\

 & H2c & 26.81 & 12.56 & 8.07 & 0.59 & 80.38 & 3.68 & 13.11 & 0.87\\

 & Tree-N & 22.84 & 16.70 & 7.11 & 0.47 & 288.35 & 2.72 & 9.53 & 0.78\\

\multirow[t]{-7}{*}{\raggedleft\arraybackslash 1000} & Tree-exp & 21.06 & 17.26 & 7.03 & 0.48 & 272.52 & 2.62 & 10.38 & 0.69\\
\cmidrule{1-10}
 & Friedman 1 & 155.81 & 139.88 & 28.70 & 3.72 & 928.84 & 40.79 & 134.97 & 4.84\\

 & Friedman 2 & 155.62 & 101.58 & 35.25 & 2.30 & 430.80 & 33.43 & 60.99 & 2.65\\

 & Friedman 3 & 106.70 & 91.72 & 34.41 & 2.36 & 330.81 & 28.47 & 73.18 & 2.72\\

 & Peak & 287.67 & 245.17 & 22.76 & 6.84 & 1914.56 & 57.96 & 123.19 & 11.80\\

 & H2c & 283.53 & 107.57 & 24.12 & 4.31 & 271.18 & 42.69 & 79.19 & 5.01\\

 & Tree-N & 106.17 & 71.78 & 22.97 & 3.19 & 753.32 & 26.88 & 74.81 & 3.86\\

\multirow[t]{-7}{*}{\raggedleft\arraybackslash 5000} & Tree-exp & 95.85 & 68.90 & 21.76 & 3.32 & 779.42 & 26.09 & 64.02 & 3.85\\
\bottomrule
\end{tabular}}
\end{table}

\begin{table}[t]

\caption{\label{table:timereal}Average computational time (in seconds) of each method over 100 times 10-fold cross-validation for each real data set. PIBF-CV: The proposed method with the cross-validation calibration, PIBF-OOB: The proposed method with the OOB calibration, RFPI: The average total running time of the five PI methods, \textit{i.e.} LM + Quant + SPI + HDR + CHDR.}
\centering
\resizebox{\linewidth}{!}{
\begin{tabular}[t]{lrrrrrrrrrr}
\toprule
Data & \textit{n} & \textit{p} & PIBF-CV & PIBF-OOB & $\widehat{PI}_\alpha$ & OOB & RFPI & SC & QRF & GRF\\
\midrule
Abalone & 4177 & 8 & 102.37 & 64.39 & 17.65 & 5.20 & 422.10 & 20.42 & 45.39 & 6.32\\

Air quality (AH) &  &  & 383.60 & 126.95 & 26.63 & 11.55 & 1256.56 & 45.45 & 126.36 & 16.07\\

Air quality (RH) & \multirow{-2}{*}{\raggedleft\arraybackslash 6941} & \multirow{-2}{*}{\raggedleft\arraybackslash 13} & 383.53 & 124.71 & 26.60 & 11.55 & 1111.16 & 46.13 & 127.02 & 15.95\\

Airfoil selfnoise & 1503 & 5 & 256.92 & 16.61 & 3.57 & 0.31 & 271.35 & 1.80 & 3.65 & 0.65\\

Ames housing & 2929 & 80 & 195.07 & 28.56 & 15.64 & 8.73 & 333.28 & 42.95 & 226.49 & 11.34\\

Auto MPG & 392 & 7 & 11.02 & 7.08 & 1.82 & 0.35 & 47.84 & 0.58 & 1.73 & 0.40\\

Boston housing& 506 & 13 & 14.77 & 8.89 & 2.32 & 0.49 & 63.20 & 1.31 & 3.63 & 0.70\\

Computer hardware & 209 & 6 & 6.53 & 3.80 & 0.93 & 0.22 & 26.49 & 0.28 & 0.98 & 0.24\\

Concrete compression & 1030 & 8 & 28.28 & 17.80 & 3.38 & 0.38 & 257.19 & 2.03 & 5.86 & 0.66\\

Concrete slump & 103 & 7 & 9.52 & 2.10 & 0.65 & 0.14 & 26.60 & 0.13 & 0.62 & 0.11\\

Energy efficiency (CL) &  &  & 85.97 & 15.93 & 2.51 & 0.34 & 341.73 & 0.94 & 2.39 & 0.52\\

Energy efficiency (HL) & \multirow{-2}{*}{\raggedleft\arraybackslash 768} & \multirow{-2}{*}{\raggedleft\arraybackslash 8} & 99.95 & 20.31 & 2.47 & 0.34 & 380.97 & 0.92 & 2.42 & 0.42\\

Servo & 167 & 4 & 12.05 & 3.99 & 0.72 & 0.15 & 53.14 & 0.12 & 0.65 & 0.15\\
\bottomrule
\end{tabular}}
\end{table}

\section{Concluding remarks}

In this paper, we have introduced an R package named \texttt{RFpredInterval}. This package implements 16 methods to build prediction intervals with random forests: a new method to build \textbf{P}rediction \textbf{I}ntervals with \textbf{B}oosted \textbf{F}orests (PIBF) and 15 different variations to produce prediction intervals with random forests proposed by \citet{roy_prediction_2020}. PIBF provides bias-corrected point predictions obtained with the one-step boosted forest and prediction intervals by using the nearest neighbour out-of-bag observations to estimate the conditional prediction error distribution.

We performed an extensive simulation study with a variety of simulated data sets and applied real data analyses to investigate the performance of the proposed method. The performance was evaluated based on the coverage level and length of the prediction intervals. We compared the performance of the proposed method to 10 existing methods for building prediction intervals with random forests. The proposed method was able to maintain the desired coverage level with both the simulated and real data sets. In terms of the PI lengths, globally, the proposed method provided the shortest prediction intervals among all methods. The conclusions drawn from the analysis of real data sets were very similar to those with the simulated data sets. This provides evidence for the reliability of the proposed method. All results obtained indicate that the proposed method can be used with confidence for a variety of regression problems.

Note that the coverage rate of prediction intervals for new observations can have several interpretations. An interesting discussion about this issue is given in \citet{mayr_prediction_2012}. In that paper, the authors presented two interpretation of coverage: sample coverage and conditional coverage. Sample coverage means that if we draw a new sample from the same population as the training sample and build PIs with a desired coverage level of $\left(1-\alpha\right)$, then the global coverage rate over this sample will be $\left(1-\alpha\right)$. The conditional coverage means that if we sample many new observations always having the same set of covariates and build PIs for them with a desired coverage level of $\left(1-\alpha\right)$, then about $\left(1-\alpha\right)100\%$ of these prediction intervals will contain the true value of the response. To hold a desired level of conditional coverage, the predictive method needs to provide the desired coverage level over the entire covariate space. On the other hand, sample coverage needs only maintain the desired coverage level over the new sample, on average. Therefore, if the conditional coverage holds, then the sample coverage also holds. In practice, predictive models are mostly evaluated with their global predictive performance. Hence, ensuring that the sample coverage level is achieved should be sufficient for most applications. The proposed calibration method with cross-validation is designed to ensure the sample coverage property. From the simulation study and real data analyses, we can see that the sample coverage is attained with the proposed calibration method.

\section{Availability}

The package is available from CRAN at \url{https://cran.r-project.org/package=RFpredInterval}. The development version is available at \url{https://github.com/calakus/RFpredInterval}.

\section{Acknowledgments}

This research was supported by the Natural Sciences and Engineering Research Council of Canada (NSERC) and by Fondation HEC Montr\'eal.

\bibliography{refs}

\newpage

\section*{Supplementary Material}

\subsection*{Data preprocessing}

In this section, we present the steps to prepare Ames Housing data set for the analysis. We use the processed version of the data set from the \texttt{AmesHousing} package.

\begin{example*}
library("AmesHousing")
AmesHousing <- make_ordinal_ames()

## Data preprocessing
# remove observations with missing values
AmesHousing <- AmesHousing[complete.cases(AmesHousing), ]

# convert the response variable in thousands
AmesHousing$Sale_Price <- AmesHousing$Sale_Price/1000

# convert the ordered factors to numeric to preserve the ordering of the factors
ord_vars <- vapply(AmesHousing, is.ordered, logical(1))
nam_ord <- names(ord_vars)[ord_vars]
AmesHousing[, nam_ord] <- data.frame(lapply(AmesHousing[, nam_ord], as.numeric))

# group together levels with less than 30 observations,
# we use the combineLevels() function from "rockchalk" package for this step
library("rockchalk")
fac_vars <- vapply(AmesHousing, is.factor, logical(1))
AmesHousing[, fac_vars] <- data.frame(
  lapply(AmesHousing[, fac_vars],
         function(x, nmin) combineLevels(x, levs = names(table(x))[table(x)<nmin], 
                                         newLabel=c("combinedLevels")),
         nmin=30)
)
\end{example*}

\subsection*{Mean PI length results for the simulated and real data sets}

Figures~\ref{figure:dgp1} to \ref{figure:dgp7} present the mean PI length of each method for each simulated data set. Similarly, figures~\ref{figure:realdata1} to \ref{figure:realdata3} show the mean PI length of each method for each real data set. The list of methods compared in the simulation study and real data analyses are as follows.

\begin{tabular}{ll}
    PIBF & Prediction intervals with boosted forests (the proposed method)\\
    $\widehat{PI}_\alpha$ & Conditional $\alpha$-level prediction interval\\
    OOB & Out-of-Bag approach\\ 
    LM & Classical method with LS splitting rule\\ 
    Quant & Quantiles with LS splitting rule\\ 
    SPI & Shortest prediction interval with LS splitting rule\\
    HDR & Highest density region with LS splitting rule\\ 
    CHDR & Contiguous highest density region with LS splitting rule\\
    SC & Split conformal\\
    QRF & Quantile regression forest\\ 
    GRF & Generalized random forest
\end{tabular}

\renewcommand{\thefigure}{S1}
\begin{figure}[htbp]
  \centering
  \includegraphics[scale=0.8]{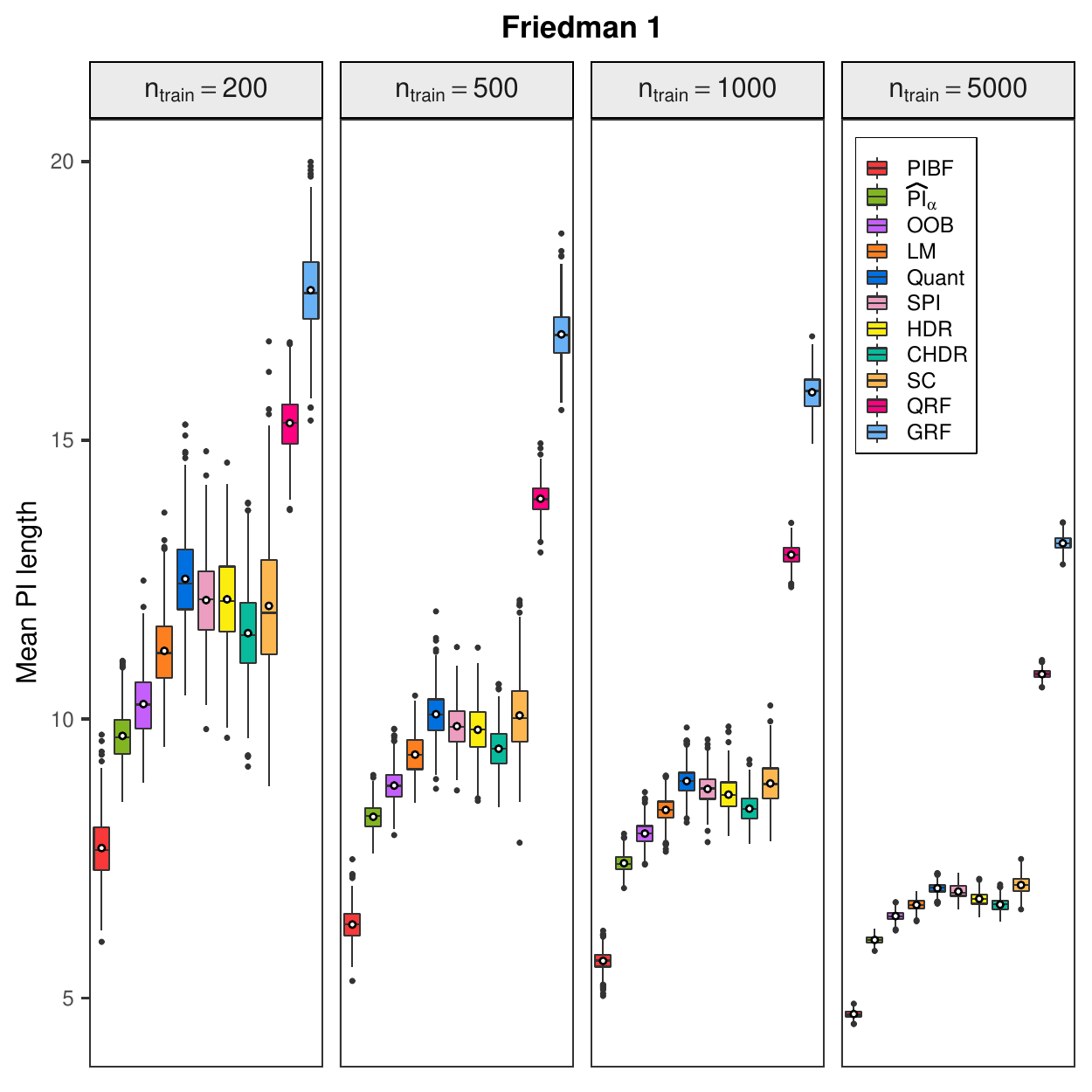}
  \caption{Distributions of the mean PI length over the test set across 500 replications for Friedman Problem 1. }
  \label{figure:dgp1}
\end{figure}

\renewcommand{\thefigure}{S2}
\begin{figure}[htbp]
  \centering
  \includegraphics[scale=0.8]{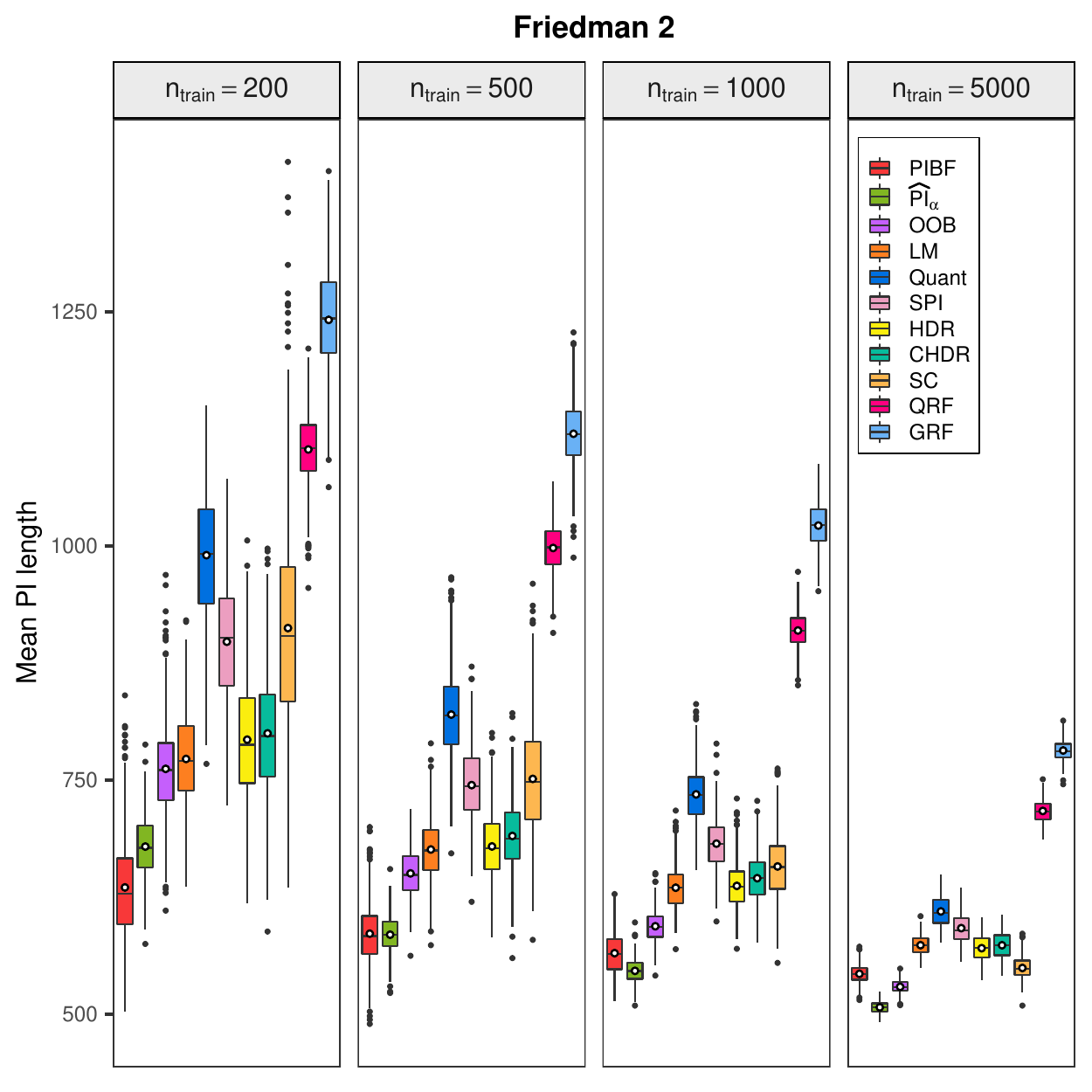}
  \caption{Distributions of the mean PI length over the test set across 500 replications for Friedman Problem 2.}
  \label{figure:dgp2}
\end{figure}

\renewcommand{\thefigure}{S3}
\begin{figure}[htbp]
  \centering
  \includegraphics[scale=0.8]{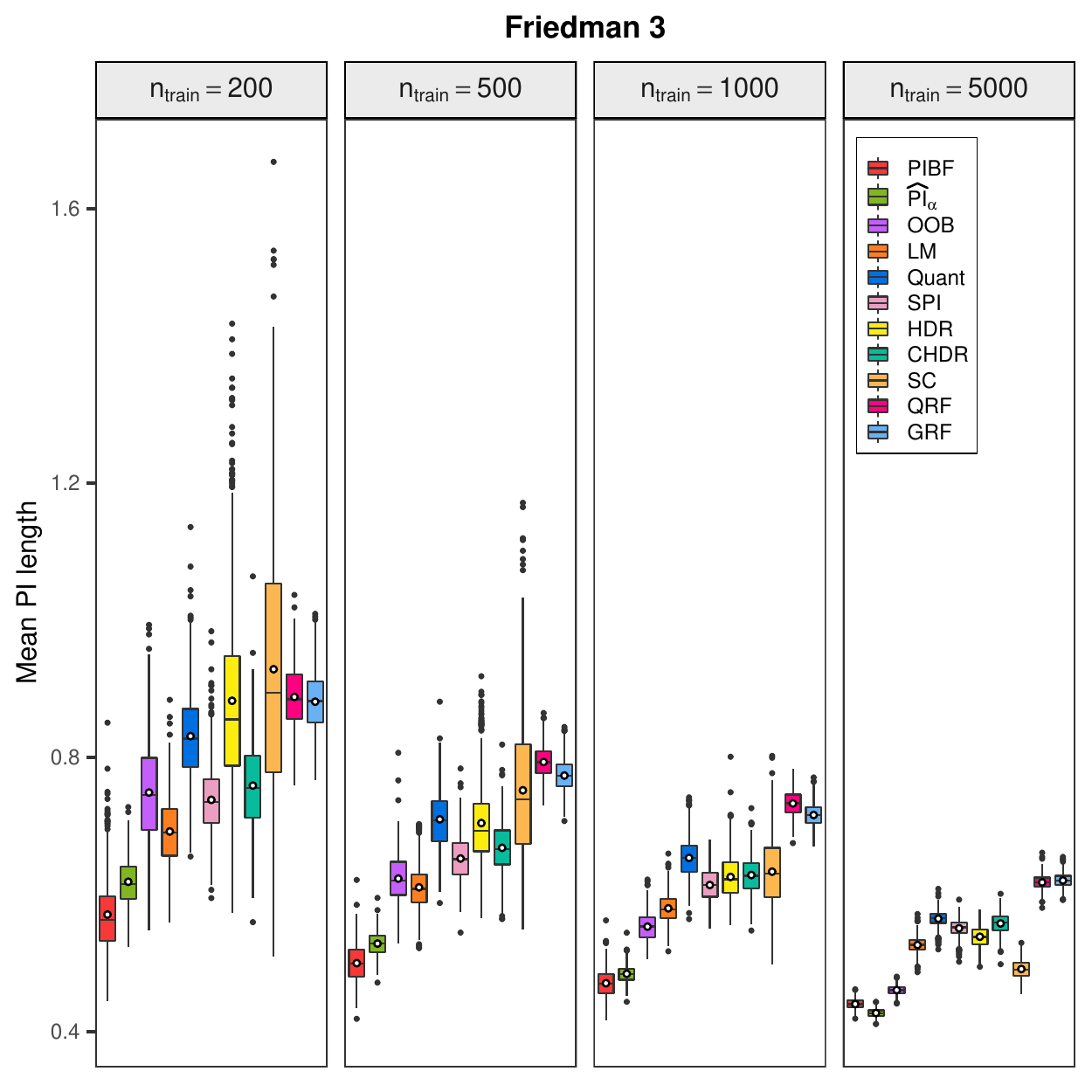}
  \caption{Distributions of the mean PI length over the test set across 500 replications for Friedman Problem 3.}
  \label{figure:dgp3}
\end{figure}

\renewcommand{\thefigure}{S4}
\begin{figure}[htbp]
  \centering
  \includegraphics[scale=0.8]{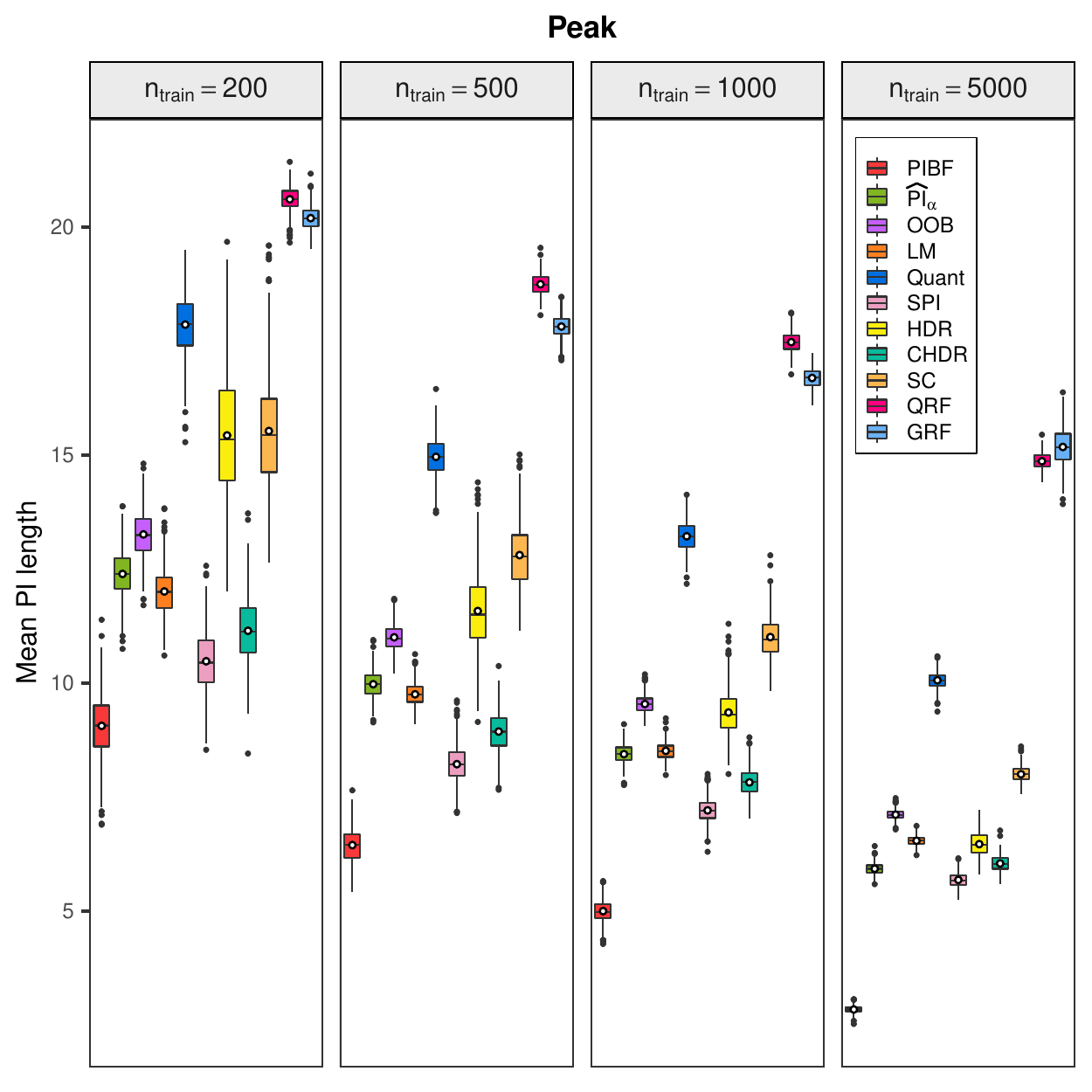}
  \caption{Distributions of the mean PI length over the test set across 500 replications for Peak Benchmark Problem.}
  \label{figure:dgp4}
\end{figure}

\renewcommand{\thefigure}{S5}
\begin{figure}[htbp]
  \centering
  \includegraphics[scale=0.8]{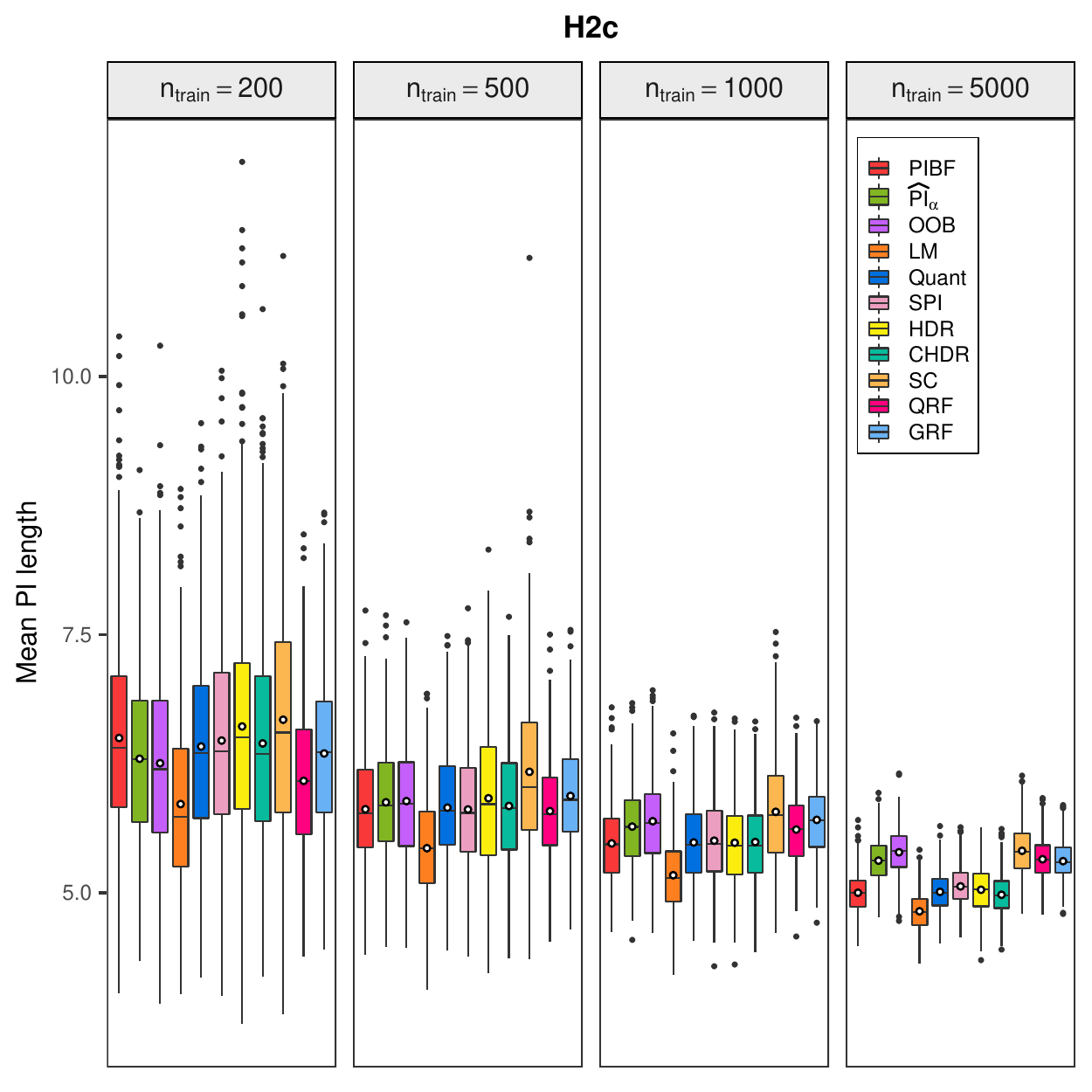}
  \caption{Distributions of the mean PI length over the test set across 500 replications for the H2c setup.}
  \label{figure:dgp5}
\end{figure}

\renewcommand{\thefigure}{S6}
\begin{figure}[htbp]
  \centering
  \includegraphics[scale=0.8]{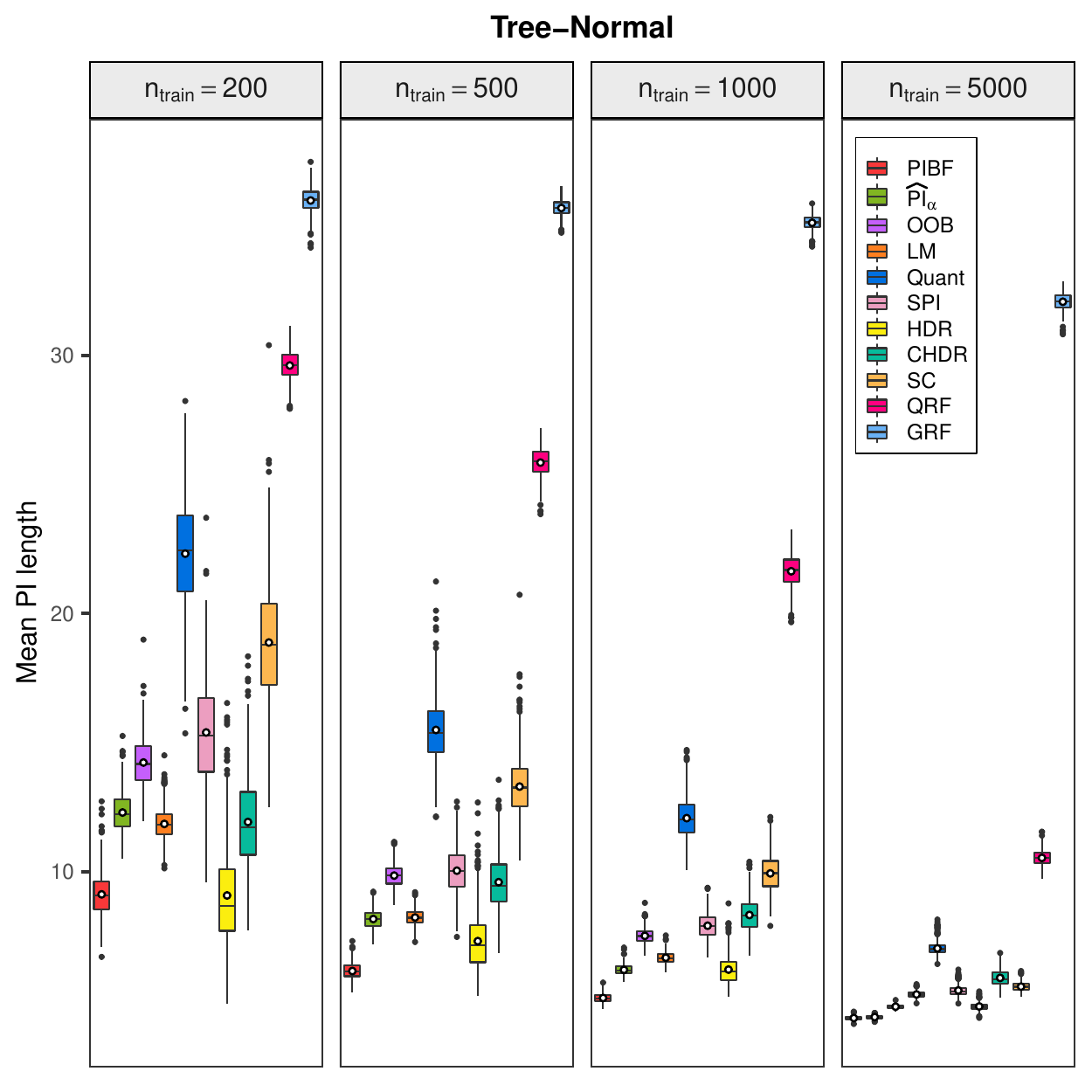}
  \caption{Distributions of the mean PI length over the test set across 500 replications for the tree-based problem with normally distributed error.}
  \label{figure:dgp6}
\end{figure}

\renewcommand{\thefigure}{S7}
\begin{figure}[htbp]
  \centering
  \includegraphics[scale=0.8]{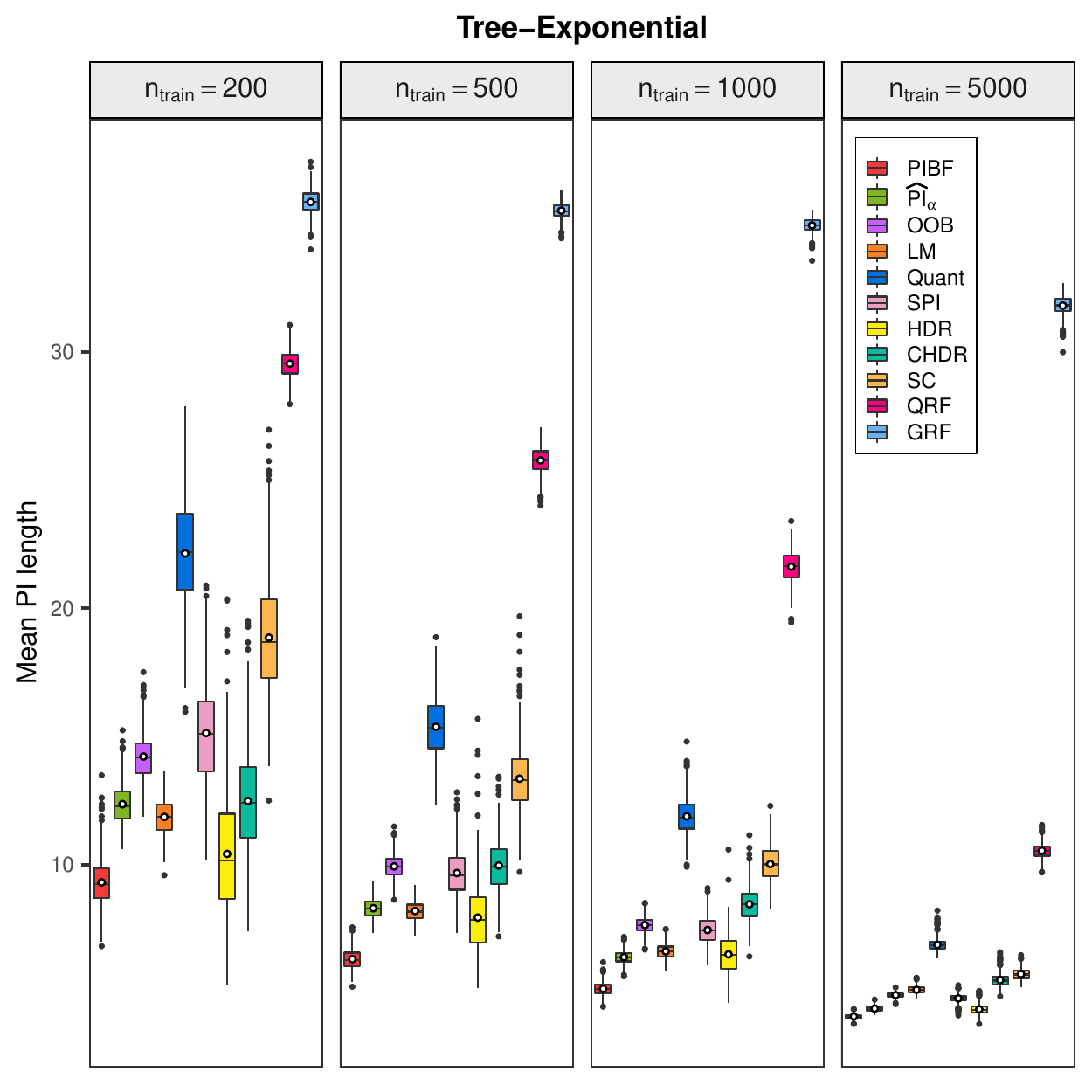}
  \caption{Distributions of the mean PI length over the test set across 500 replications for the tree-based problem with exponentially distributed error.}
  \label{figure:dgp7}
\end{figure}

\renewcommand{\thefigure}{S8}
\begin{figure}[htbp]
  \centering
  \includegraphics[scale=0.8]{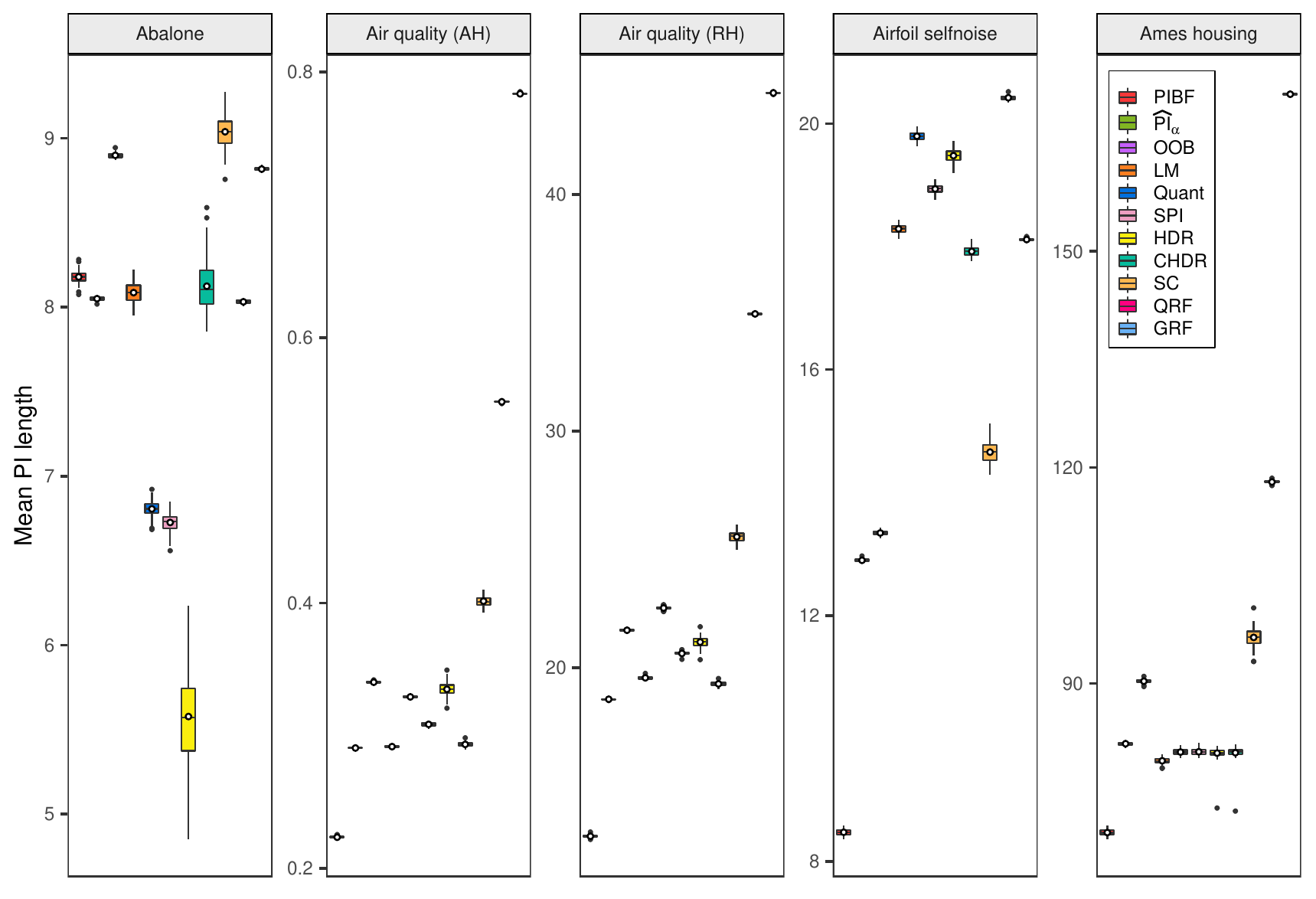}
  \caption{Distributions of the mean PI length across 100 repetitions for Abalone, Air quality with absolute and relative humidity, Airfoil selfnoise, and Ames housing data sets.}
  \label{figure:realdata1}
\end{figure}

\renewcommand{\thefigure}{S9}
\begin{figure}[htbp]
  \centering
  \includegraphics[scale=0.8]{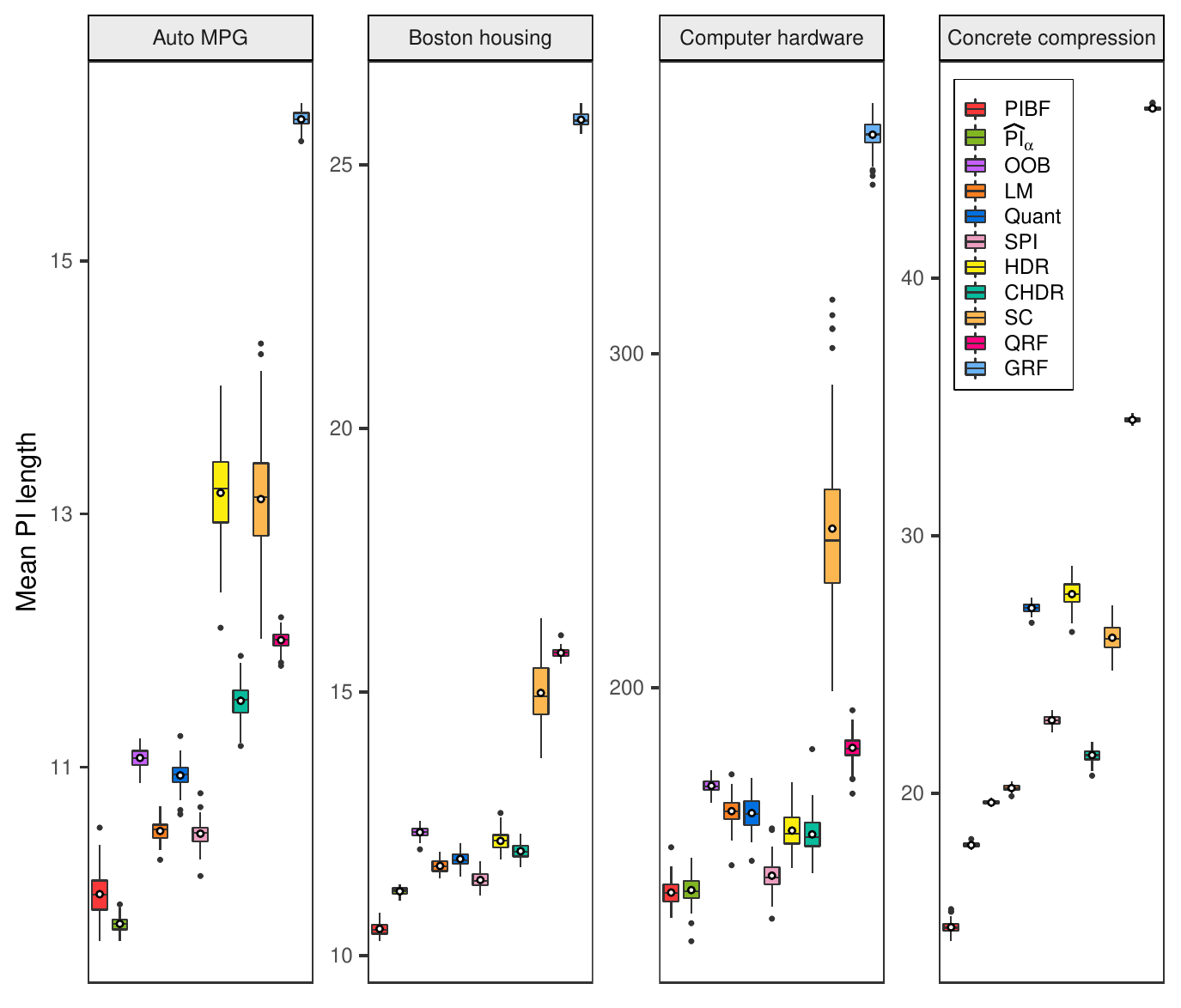}
  \caption{Distributions of the mean PI length across 100 repetitions for Auto MPG, Boston housing, Computer hardware, and Concrete compression data sets.}
  \label{figure:realdata2}
\end{figure}

\renewcommand{\thefigure}{S10}
\begin{figure}[htbp]
  \centering
  \includegraphics[scale=0.8]{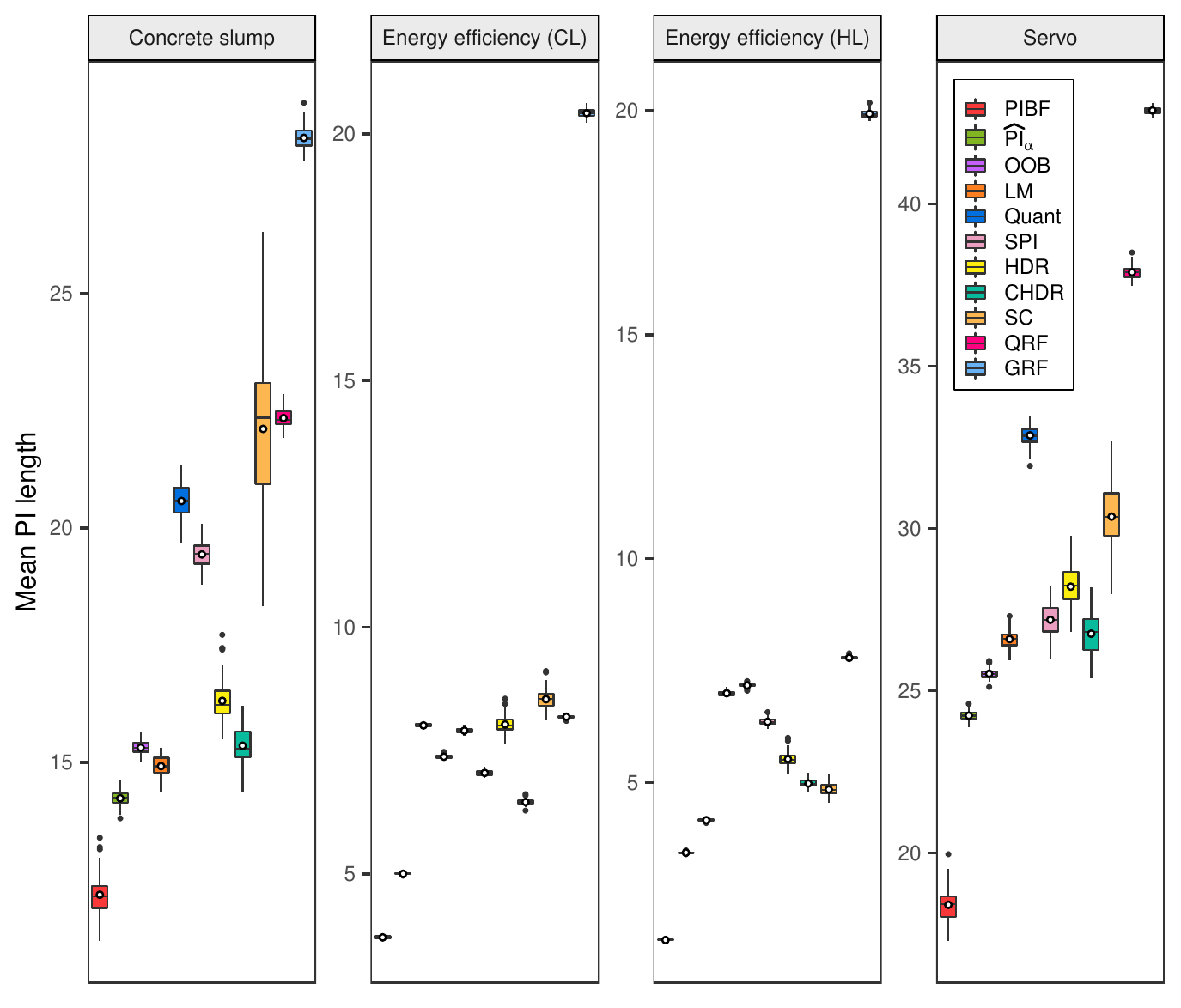}
  \caption{Distributions of the mean PI length across 100 repetitions for Concrete slump, Energy efficiency with cooling and heating load, and Servo data sets.}
  \label{figure:realdata3}
\end{figure}

\end{document}